\pdfoutput=1
\documentclass[10pt,twocolumn,letterpaper]{article}
\usepackage[pagenumbers]{wacv}

\usepackage{booktabs}
\usepackage{multirow}
\usepackage{longtable}
\usepackage{tabularx}
\usepackage{tikz}
\usetikzlibrary{arrows.meta,positioning,fit,calc,backgrounds}
\usepackage{fvextra}
\fvset{formatcom=\fontencoding{T1}\selectfont}
\usepackage{fontawesome5}
\usepackage{colortbl}
\definecolor{zebragray}{gray}{0.945}
\newcommand{\zebra}{\rowcolor{zebragray}}
\newcommand{\hd}[1]{\textbf{#1}}
\DeclareRobustCommand{\cnum}[1]{\tikz[baseline=(c.base)]\node[circle,fill=black!75,text=white,inner sep=0pt,minimum size=1.9ex,font=\sffamily\tiny\bfseries](c){#1};}
\DeclareUnicodeCharacter{21D2}{\ensuremath{\Rightarrow}}
\DeclareUnicodeCharacter{2264}{\ensuremath{\leq}}
\DeclareUnicodeCharacter{2265}{\ensuremath{\geq}}

\definecolor{wacvblue}{rgb}{0.21,0.49,0.74}
\usepackage[breaklinks,colorlinks,allcolors=wacvblue]{hyperref}

\def\wacvPaperID{*****}
\def\confName{WACV}
\def\confYear{2027}

\newcommand{\datarepo}{\href{https://github.com/anoubhav/ad-review-and-edit}{\mbox{\texttt{github.com/anoubhav/ad-review-and-edit}}}}

\title{Multimodal Judges Flag, Humans Decide: A Review-and-Edit\\Framework for AI-Generated Product Video Ads}

\author{Anoubhav Agarwaal \quad Sukesh Perla\textsuperscript{*} \quad Karthikeyan Mohanraj\textsuperscript{*} \quad Susmita Ghose\\
Anoki Inc.\\
{\tt\small \{anoubhav, sukeshp, karthikeyanm, susmita\}@anoki.tv}
}

\hypersetup{pdftitle={Multimodal Judges Flag, Humans Decide: A Review-and-Edit Framework for AI-Generated Product Video Ads},pdfauthor={Anoubhav Agarwaal, Sukesh Perla, Karthikeyan Mohanraj, Susmita Ghose},pdfkeywords={AI-generated video, video ads, video quality assessment, VLM-as-a-judge, LLM-as-a-judge, multimodal judges, human-in-the-loop review, product fidelity, mispronunciation detection, automation bias}}

\usepackage{stfloats}
\begin{document}
\crefname{appendix}{Supp.}{Supp.}\Crefname{appendix}{Supp.}{Supp.}
\raggedbottom
\twocolumn[{\renewcommand\twocolumn[1][]{#1}\maketitle
\begin{center}
\definecolor{lpModel}{HTML}{2A78D6}%
\definecolor{lpPeople}{HTML}{EB6834}%
\definecolor{lpTool}{HTML}{2F9E6E}%
\colorlet{lpSec}{black!72}%
\providecommand{\pmk}{\textcolor{lpPeople}{\rule[0.56mm]{1.5mm}{0.38mm}\kern-0.94mm\rule{0.38mm}{1.5mm}\kern0.56mm}}%
\begin{tikzpicture}[x=1cm, y=1cm, font=\sffamily\scriptsize\frenchspacing, line width=0.6pt,
  arr/.style={draw=black!50, -{Latex[length=1.5mm, width=1.3mm]}, rounded corners=3pt},
  card/.style={draw=#1!70, fill=#1!7, rounded corners=3pt, line width=0.6pt, inner xsep=1.8mm, inner ysep=1.4mm, align=center},
  lens/.style={draw=black!25, fill=white, rounded corners=3pt, line width=0.5pt},
  lab/.style={text=lpSec, inner sep=1pt},
  jm/.style={rectangle, fill=#1, draw=none, inner sep=0pt, minimum size=1.3mm},%
  pm/.style={inner sep=0pt, text=lpPeople},%
  vt/.style={inner sep=0pt, text=lpPeople, font=\footnotesize},%
  tl/.style={font=\sffamily\scriptsize\bfseries, text=lpSec, inner sep=0pt, anchor=base},
  ql/.style={anchor=base west, inner sep=0pt},
  row/.style={anchor=base west, inner sep=0pt},
  trk/.style={draw=black!25, fill=black!6, rounded corners=1.2pt},
    lead/.style={draw=#1, line width=0.5pt, rounded corners=2pt},
]
\def\FL{2.28}\def\FR{13.59}\def\FT{-0.15}\def\FB{-4.36}
\def\TX{4.56}\def\SP{1.05}\def\SW{0.98}\def\TE{8.69}%
\def\CL{8.95}\def\CR{13.45}%
\def\yA{-0.85}%
\node[card=lpModel, align=left] (gen) at (0.98,\yA)
  {\textbf{Generate}\\[1pt]{\color{lpSec}an ad from}\\{\color{lpSec}product data}};
\draw[black!30, fill=black!2, rounded corners=4pt] (\FL,\FT) rectangle (\FR,\FB);
\draw[arr] (gen.east) -- (\FL,\yA);
\node[anchor=west, fill=white, inner xsep=2pt, inner ysep=1pt] at (\FL+0.15,\FT)
  {\textbf{Review}\enspace{\color{lpSec}six axes}};
\node[anchor=west, fill=white, inner xsep=2pt, inner ysep=1pt, text=lpSec] at (\CL+0.10,\FT) {two lenses on the same ad};
\node[ql] at (2.44,-0.80) {\makebox[2.3mm][l]{\color{lpSec}\bfseries 1}Product};
\node[ql] at (2.44,-1.10) {\makebox[2.3mm][l]{\color{lpSec}\bfseries 2}Scene quality};
\node[ql] at (2.44,-2.16) {\makebox[2.3mm][l]{\color{lpSec}\bfseries 3}Pronunciation};
\node[ql] at (2.44,-2.46) {\makebox[2.3mm][l]{\color{lpSec}\bfseries 4}Script};
\node[ql] at (2.44,-2.98) {\makebox[2.3mm][l]{\color{lpSec}\bfseries 5}Sound};
\node[ql] at (2.44,-3.45) {\makebox[2.3mm][l]{\color{lpSec}\bfseries 6}Captions};
\foreach \i in {1,...,4}{
  \fill[black!10, rounded corners=1.2pt] ({\TX+(\i-1)*\SP},-0.45) rectangle ++(\SW,-0.90);
  \node[tl] at ({\TX+(\i-1)*\SP+\SW/2},-0.70) {scene \i};
}
\def\CT{-3.30}\def\CM{-3.405}
\foreach \i in {1,...,4}{
  \draw[black!30, fill=white, rounded corners=1pt] ({\TX+(\i-1)*\SP+0.05},\CT) rectangle ++(0.60,-0.21);
  \draw[black!35, line width=1pt] ({\TX+(\i-1)*\SP+0.14},\CM) -- ++(0.42,0);
}
\foreach \a/\b/\n in {4.56/5.74/1, 5.81/6.80/2, 6.87/7.80/3, 7.87/8.69/4}{
  \draw[trk] (\a,-1.98) rectangle (\b,-2.58);
  \node[text=lpSec, inner sep=0pt] at ({(\a+\b)/2},-2.28) {line \n};}
\draw[trk] (\TX,-2.78) rectangle (\TE,-3.09);
\node[text=lpSec, inner sep=0pt] at ({(\TX+\TE)/2},-2.935) {whole ad};
\def\MX{0.24}\def\RX{0.44}%
\def\VU{12.95}\def\VD{13.22}%
\draw[lens] (\CL,-0.37) rectangle (\CR,-2.10);
\node[row] at (\CL+0.15,-0.63) {{\color{lpModel}\faRobot\enspace\textbf{Judges}}\enspace{\color{lpSec}(AI models)}};
\node[row] at (\CL+0.15,-0.93) {place findings on a scene, a line or};
\node[row] at (\CL+0.15,-1.23) {the whole ad, inside the questions};
\node[jm=lpModel] (jk) at (\CL+\MX,-1.49) {};
\node[row] at (\CL+\RX,-1.55) {scene 3: product logo garbled};
\node[jm=lpModel] (jk2) at (\CL+\MX,-1.82) {};
\node[row] at (\CL+\RX,-1.88) {line 2: claim not in the listing};
\node[vt] (vu) at (\VU,-1.49) {\faThumbsUp};
\node[vt] (vd) at (\VD,-1.82) {\faThumbsDown};
\draw[lens] (\CL,-2.22) rectangle (\CR,-4.24);
\node[row] at (\CL+0.15,-2.48) {{\color{lpPeople}\faUser\enspace\textbf{Reviewer}}\enspace{\color{lpSec}(a person)}};
\node[row] at (\CL+0.15,-2.78) {answers all six questions};
\node[vt] at (\CL+\MX,-3.04) {\faThumbsUp};
\node[row] (ra) at (\CL+\RX,-3.10) {accepts the logo finding};
\node[vt] at (\CL+\MX,-3.37) {\faThumbsDown};
\node[row] (rr) at (\CL+\RX,-3.43) {rejects the line 2 finding};
\node[row] at (\CL+0.15,-3.74) {adds what the judges missed:};
\node[pm] (rk) at (\CL+\MX,-3.97) {\pmk};
\node[row] at (\CL+\RX,-4.03) {scene 3: caption covers the logo};
\draw[lead=lpPeople!70] ($(ra.base east)+(1.2mm,0.6mm)$) -| (vu.south);
\draw[lead=lpPeople!70] ($(rr.base east)+(1.2mm,0.6mm)$) -| (vd.south);
\node[jm=lpModel] (p1) at ({\TX+2*\SP+0.68},-1.08) {};
\draw[lead=lpModel!70] (jk.west) -| (p1.south);
\node[jm=lpModel] (p3) at (6.66,-2.10) {};%
\draw[lead=lpModel!70] (jk2.west) -| (p3.north);
\draw[lpPeople, line width=0.8pt, rounded corners=1pt] ({\TX+2*\SP+0.05},\CT) rectangle ++(0.60,-0.21);
\node[pm] (p2) at ({\TX+2*\SP+0.82},\CM) {\pmk};
\draw[lead=lpPeople!70] (rk.west) -| (p2.south);
\node[draw=black!50, fill=black!5, rounded corners=3pt, inner xsep=1.6mm, inner ysep=1.1mm, anchor=west] (dec) at (\FR+0.40,\yA)
  {Decide};
\draw[arr] (\FR,\yA) -- (dec.west);
\node[text=black!70, font=\normalsize, inner sep=0pt] (ok) at (16.92,\yA) {\faCheckCircle};
\draw[arr] (dec.east) -- (ok.west) node[midway, above=1pt, lab] {approve};
\node[lab, anchor=north] at (ok.south) {delivered};
\def\EL{14.00}\def\ER{17.35}\def\ET{-2.10}\def\EB{-3.56}
\draw[lpTool!70, fill=lpTool!7, rounded corners=3pt] (\EL,\ET) rectangle (\ER,\EB);
\draw[arr, draw=lpTool] (dec.south) -- (dec.south |- 0,\ET) node[midway, right=1pt, lab, text=lpTool!80!black] {needs edits};
\node[row] at (\EL+0.15,-2.44) {\textbf{Edit}};
\node[row] at (\EL+0.15,-2.72) {scene 3 redone,};
\node[row] at (\EL+0.15,-3.00) {its caption moved};
\foreach \i/\t in {1/1,2/2,3/new,4/4}{
  \ifnum\i=3 \def\fc{lpTool}\def\tc{white}\else\def\fc{black!13}\def\tc{lpSec}\fi
  \fill[\fc, rounded corners=1pt] ({\EL+0.15+(\i-1)*0.70},-3.12) rectangle ++(0.64,-0.32);
  \node[text=\tc, inner sep=0pt, font=\sffamily\scriptsize\bfseries] at ({\EL+0.15+(\i-1)*0.70+0.32},-3.28) {\t};}
\draw[arr, draw=lpTool, dash pattern=on 2.2pt off 1.6pt] (\EL+0.40,\EB) -- (\EL+0.40,-4.62) -- (8.20,-4.62) -- (8.20,\FB);
\node[lab, anchor=west, text=lpTool!80!black] at (\EL+0.52,-4.05) {edited ad, reviewed again};
\end{tikzpicture}


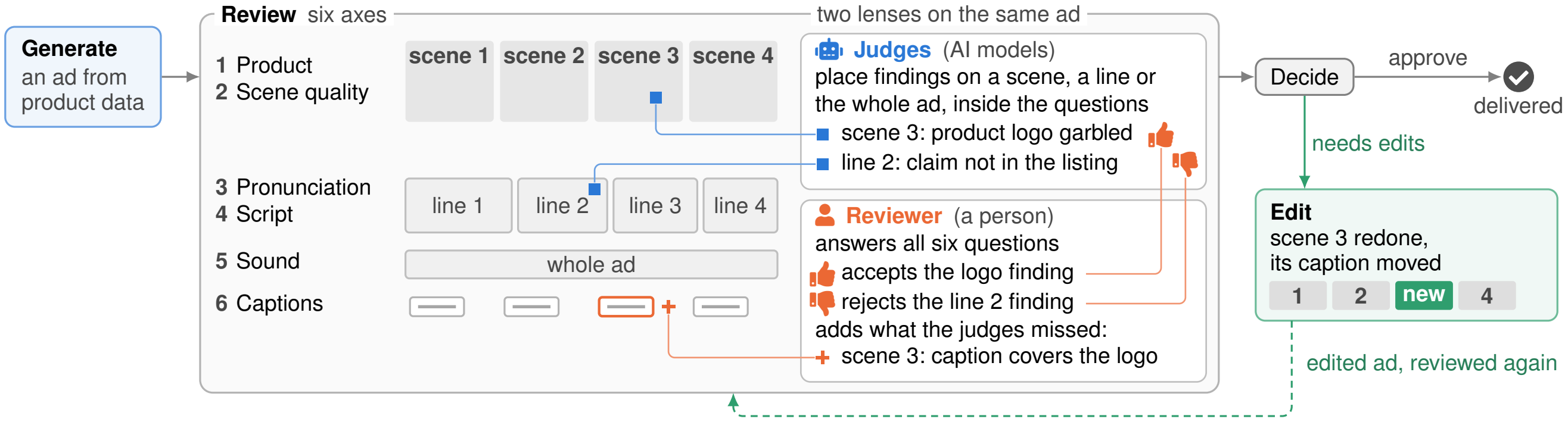
\captionof{figure}{A review-and-edit framework for AI-generated product video ads, built to run at catalog scale. AI judges and the reviewer work on the same six axes, each shown beside the part of the ad it judges (\cref{tab:axes}): the judges (blue) flag problems, and the reviewer (orange) answers one question per axis, votes (thumb) on each judge finding (square) and adds what the judges missed (plus). At Decide, the ad is approved, sent to Edit (as here) or rejected (\cref{sec:axes}). Edit fixes only what the review found (here, redoing scene 3 and moving its caption), keeps the rest of the ad, and sends the edited ad back to review.}
\label{fig:loop}
\end{center}
\vspace{0.5ex}
}]
{\let\thefootnote\relax\footnotetext{\textsuperscript{*}Equal contribution.}}
\begin{abstract}
Generative pipelines can now turn every product page of an e-commerce catalog into a short video ad, but deciding whether such an ad is usable is not one good-or-bad call. We describe how we operationalized the review and editing of these ads so that both can keep pace with generation: reviewers go through each ad scene by scene and answer six questions, on the product, scene quality, pronunciation, script, sound and captions; judges built from vision-language, speech and language models file each finding under the matching question, and reviewers accept or reject each finding and add what the judges missed. Each confirmed issue then opens an edit tool matched to it. The framework is flexible: its questions and the routing from a finding to its fix do not depend on the video, image, voice or text models that generate the ad, and the six questions stayed the same as we changed our generation models. We apply it to our own pipeline. Across 120 of its ads, reviewed with judge findings on screen, the visuals were the main problem: reviewers marked the product wrong on 60\% of scenes, and script, caption or sound problems on under 10\% each. On 60 of these ads, they marked the product wrong on 58\% of scenes, far more than the 26\% marked by two reviewers who later saw the ads with the judges hidden. One reviewer who saw these ads both ways, two weeks apart, marked the product wrong on 57\% of scenes with judge findings and 19\% without.{} On 120 blind-labeled scenes, a product judge that watches the clip with the catalog photos reached AUC 0.84, while video-quality and photo-similarity scores were close to chance. We tune the judges for recall, since a missed defect costs more than a false flag: the product judge catches nearly all defects but also flags many clean scenes, so reviewers decide; together, the judges widen what reviewers mark and reviewers add what the judges miss. Without judge findings, one reviewer's marking faded over a session; with them, it held steady. To help others replicate this review in their own pipelines, we release a review-portal template and our full review data at \datarepo: 120 ads and 562 scenes with the judge findings and human marks, plus 120 blind-labeled scenes with baseline scores.
\end{abstract}

\section{Introduction}
\label{sec:intro}

Short video ads are increasingly generated: models cut them from product footage, generate them from a product's selling points and serve them in production~\cite{zhou2026autocut,liu2026kdcvg,bu2026agenticgen}. E-commerce has long checked its catalog automatically, extracting and correcting product attributes~\cite{zheng2018opentag,dong2020autoknow,wang2020metabridge} and verifying product images~\cite{lin2021pam,gandhi2020scalable,khandelwal2023mxt}, and generated ad images are already scored and inspected at scale~\cite{yang2024cg4ctr,chen2025caig,du2024reliablead}. Video is the next step. Our pipeline turns a product's own data (its listing, specifications and catalog photos) into a 15-second ad with a script, narration, four or five generated scenes, captions and an end card, so it can make an ad for every catalog product. This paper is about what makes such ads usable at that scale: reviewing and editing them, taking the generator as given.

That review is hard, for reviewers and models alike. Every product has its own details, and an ad is usable only if it stays true to them. Yet generative models invent plausible details, in the pictures and in the words: an extra pocket on a pair of pants, a suitcase wheel in the wrong place, a script that calls a suitcase `unbreakable' when its listing does not. Reviewers can miss such a defect but agree on it once it is pointed out: on scenes more than one reviewer saw, 90\% of the product problems the judge had not flagged were marked by a single reviewer, against 12\% of those it had flagged. Problems are also local and of different kinds: one scene garbles the label while the next is fine, the narration mispronounces the brand, a caption covers a face. Calling the ad good or bad hides where each problem is, and discards ads that redoing a scene or a line could save.

Neither people nor judges can carry this review alone. Judges are hard to build: generic video-quality and photo-similarity scores miss product errors, many speech-to-text models write down the word the script expects and so hide the mispronunciation they should catch, and a model's pick of caption placement changes from run to run (\cref{sec:judges}). Even the best product judge flags many clean scenes. People struggle too: they miss defects no one points out, yet with findings on screen some of their marks may only follow the judges (\cref{sec:portal}). We therefore built a single review that judges and reviewers share (\cref{fig:loop}).

We report how review and editing work in a commercial pipeline generating e-commerce video ads at catalog scale:
\begin{enumerate}[leftmargin=*,itemsep=1pt,topsep=2pt]
\item \textbf{We frame ad review as six questions asked scene by scene} (\cref{sec:axes}). Product, scene quality, pronunciation, script, sound and captions are each asked at the part of the ad they concern, with written boundaries between them (\cref{tab:axes}), so the same review works for every ad in a catalog and every answer points to what to fix.
\item \textbf{We review each ad through two lenses, AI judges and human reviewers, and measure what each sees} (\cref{sec:portal}). Reviewers with judges mark more than reviewers alone: with judge findings on screen, reviewers marked the product wrong on far more scenes than two reviewers who later saw the same ads with the judges hidden. In exploratory checks, hidden reviews that took as long still marked less, and one reviewer's attention faded over a session with judges hidden, while their marks on the same ads with findings held steady (\cref{fig:ww}b).
\item \textbf{We release a review-portal template and a dataset from 120 generated ads.} The template\footnote{In the \href{https://github.com/anoubhav/ad-review-and-edit/tree/main/review}{\texttt{review/}} folder of the repository.} is a minimal, single-file version of our review page, the six questions scene by scene with stored judge findings shown or hidden, so others can try the review and adapt it to their own ad-generation pipelines. The dataset holds the ads' 562 scenes with every reviewer's marks and the judges' findings with the reviewers' votes, and 120 scenes from other ads, labeled before any judge existed (blind labels), with every baseline's score (catalog photos as hashes only). Both are at \datarepo.
\item \textbf{We build a judge for each question that sees what a reviewer sees} (\cref{sec:judges}). On these blind labels, the product judge, which watches the clip with every catalog photo, reached AUC 0.84, while generic video-quality and photo-similarity scores reached 0.48 to 0.57; \cref{tab:experiments} ties each design choice to its labels.
\item \textbf{We turn each confirmed issue into a targeted edit} (\cref{sec:fixing}). An issue opens the edit tool for its own scene, line or caption, which changes the smallest unit at fault and leaves the rest of the ad unchanged; the edited ad returns to review. In tests outside the full loop, Repair, which fixes a scene in place with a video model, fully corrected garbled product text in 18 of 26 rated scenes, and regenerating only the flagged script lines with a language model left most rewritten scripts clean.
\end{enumerate}
The full rubric (\cref{tab:rubric}), baselines and label sets are in the supplement.

\section{Related Work}
\label{sec:related}

\textbf{Judging generated images and video.}\enspace{}Judges built on multimodal language models score generated content, with known biases relative to people~\cite{zheng2023judging,chen2024mllmjudge}. VIEScore and Q-Bench test models as judges of conditional synthesis and low-level quality~\cite{ku2024viescore,wu2024qbench}, VBench scores generated video on many dimensions~\cite{huang2024vbench}, and EvalCrafter, T2VQA-DB and VideoScore fit video-quality scores to human ratings~\cite{liu2024evalcrafter,kou2024t2vqa,he2024videoscore}; these rate a whole clip, while our judges name a defect, its scene and its evidence. Specialist models~\cite{do2026specialists} and scorers learned from human preferences~\cite{kas2026brokenlimbs} strengthen judges. Product fidelity is usually measured with embedding or perceptual similarity~\cite{oquab2024dinov2,radford2021clip,zhang2018lpips,fu2023dreamsim} or product-consistency models~\cite{liang2025hfpc,du2024reliablead}, and RichHF localizes faults from human feedback~\cite{liang2024richhf}. Our judges each check one part of an ad; we test them against blind labels and reviewers' marks.

\begin{table*}[t]
\centering\footnotesize
\caption{The six review axes, one per row. Columns: the unit an answer points to (a line is the narration spoken over a scene); the answers, with the secondary options that name the kind of problem; and what the judge sees. AI issue: a defect typical of generated video, such as melted detail or a duplicated part. Each kind of problem has its own fix (\cref{tab:fixes}); \cref{app:gallery} shows an example of each but the music.}
\label{tab:axes}
\setlength{\tabcolsep}{3.5pt}
\begin{tabularx}{\textwidth}{@{}l l >{\raggedright\arraybackslash}X >{\raggedright\arraybackslash}p{7.2cm}@{}}
\toprule
\hd{Axis} & \hd{Unit} & \hd{Answer and secondary options} & \hd{Judge: what it sees} \\
\midrule
\textbf{Product} & scene & right / wrong; wrong product; shape, color, logo or markings wrong; AI issue on the product; product text not intact (garbled; low: readable, high: gibberish) & a vision-language model watches the clip with all catalog photos and an AI-made product sheet, citing time, place and photo \\
\zebra \textbf{Scene quality} & scene & everything in the scene but the product: keep / swap / no opinion; AI issue; repetitive, dull or crowded; world or continuity breaks; impossible motion; weak ending & two models: one watches the clips in order, the other sees four frames per scene \\
\textbf{Pronunciation} & line & sounds right / mispronounced, with the word & a speech-to-text model transcribes the line's narration, keeping mispronunciations; a text model checks each mismatch against the script \\
\zebra \textbf{Script} & line & good / bad; unsupported claim; awkward phrasing & a language model checks each claim against the listing and listing-photo text, quoting the words \\
\textbf{Sound} & ad & right / issues: the voice, the music & one audio-model verdict on the whole mix, with no scene or time \\
\zebra \textbf{Captions} & caption & placement: correct / can be better; relevance: correct / issues & judges watch the finished ad (run after the review round, \cref{sec:portal}); at generation a model places captions off faces and logos, and clutter and contrast are measured \\
\bottomrule
\end{tabularx}
\end{table*}

\textbf{Generated ads and production pipelines.}\enspace{}Generated image ads are already selected or optimized with click-trained reward models~\cite{yang2024cg4ctr,chen2025caig} and inspected for defects and regenerated~\cite{du2024reliablead,liang2025hfpc}. Video-ad systems cut ads from footage, generate them from a product's selling points or serve them in production~\cite{zhou2026autocut,liu2026kdcvg,bu2026agenticgen}; AdsQA, VideoAds and E-VAds benchmark how well models understand ads~\cite{long2025adsqa,zhang2025videoads,evads2026}, and editors' critiques of generated video ads yield six dimensions of editing quality~\cite{law2026editors}. Pipelines already pair generation with critique and local regeneration~\cite{das2026genflow,long2025vista,atighehchian2026prompt,xu2026commercevibe}, and an automatic quality stage with modest precision can roughly halve the cost per accepted generated image~\cite{giro2026cost}. These works score a whole ad, optimize it for clicks or study editing craft; we check every ad a catalog pipeline makes, scene by scene against the product's own data, and send each issue a reviewer confirms to an edit tool.

\textbf{People working with automated checks.}\enspace{}Content moderation pairs classifiers with human review~\cite{gorwa2020moderation}; people over-rely on automated advice~\cite{parasuraman2010complacency,goddard2012automation}, and human--AI teams do not reliably beat the better of the two~\cite{bansal2021whole}. Visible model suggestions pull human labels toward the model~\cite{schroeder2025humanloop}, and rater training changes the quality of subjective labels~\cite{rahul2026subjective}. Our checks also draw on claim verification against evidence~\cite{thorne2018fever} and mispronunciation detection~\cite{witt2000phone}; mean-opinion-score predictors score speech noise and naturalness~\cite{reddy2021dnsmos,saeki2022utmos}, while our pronunciation question asks if a given word was said right.

\section{Method: Six Axes, Scene by Scene}
\label{sec:axes}

\begin{figure*}[t]
\centering
{\def\stripframes{figures/strip/}\input{figures/fig_example.tex}}
\caption{One generated 15\,s ad (\href{https://github.com/anoubhav/ad-review-and-edit/blob/main/example/ad-2438b25fb3.mp4}{video} in our release) as judges and reviewers reviewed it, one row per question. Reviewers accepted the judges' product findings and rejected nearly all of their pronunciation findings. They also marked the S3 and S4 captions and added two problems the judges missed in S2: an impossible motion in the scene and an unsupported claim in its line.}
\label{fig:example}
\end{figure*}

This section defines the six review questions, what a reviewer sees with and without judges, and how reviews become verdicts.

\paragraph{Six axes.}
We review an ad on six axes (\cref{tab:axes}), above all the product, which must match its catalog photos. Product and scene quality are both about the visuals; scene quality covers everything in the scene but the product. They are asked separately because they tend to need different fixes: a good scene with a flawed product can often be fixed in place, by Repair or a clip edit, while a weak scene is usually swapped or made anew (\cref{tab:fixes}). We ask one question per axis, in two parts for captions (placement and relevance; \cref{tab:axes}); \cref{fig:example} shows them on one ad of our review round (\cref{sec:portal}), and \cref{app:gallery} shows an example of every kind of problem but the music. A ``wrong'' answer opens a picker for the scenes, lines or captions at fault. Reviewers and judges use the same questions, so a judge's finding, a reviewer's mark and a fix speak the same language: the same scene and the same kind of problem. Terms: an ad is a sequence of four or five generated scenes. A caption is the overlay text on the picture; ads on e-commerce catalogs and marketplaces often play muted, so captions carry the message. A finding is one problem a judge raises, and a flag is a scene, line or caption with a finding. A mark is one reviewer's answer that something is wrong, and a review is one reviewer's answers on one ad.

\paragraph{How a review runs.}
The reviewer watches the whole ad first and can replay it or any scene as often as needed. Most answers point to one scene, line or caption, but some need the whole ad as context: a repeated scene, a background that jumps between scenes or a weak ending only shows when the scenes are seen together, and sound is judged over the full mix.

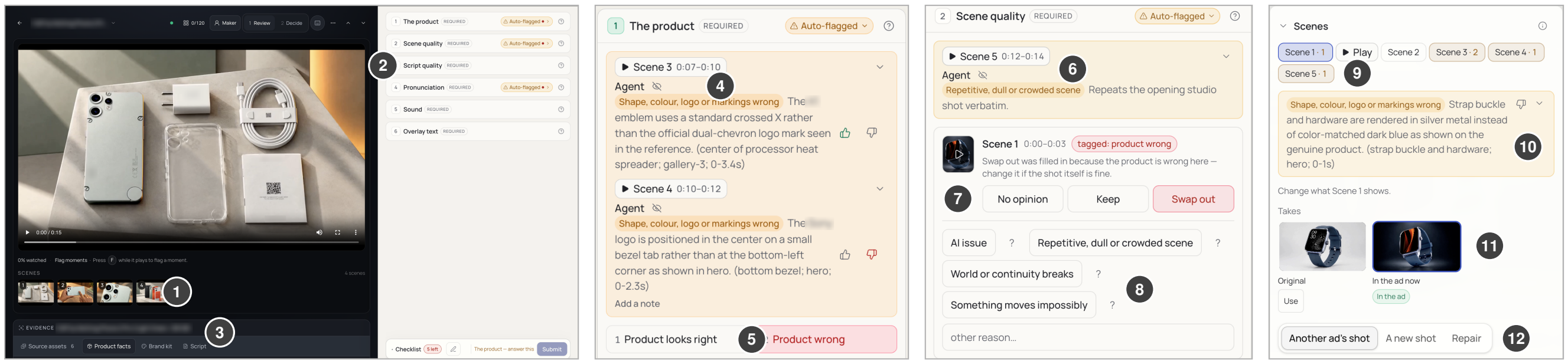
\begin{figure*}[t]
\centering
\newlength{\uiH}\setlength{\uiH}{3.95cm}
\newcommand{\uipanel}[3][]{\begin{tikzpicture}
  \node[inner sep=0, anchor=south west] (img) {\includegraphics[height=\uiH]{#2}};
  \draw[black!30] (img.south west) rectangle (img.north east);
  \begin{scope}[x={(img.south east)}, y={(img.north west)}]
  \ifx\relax#1\relax\else\foreach \ax/\ay/\bx/\by in {#1}{\draw[black!60, line width=0.4pt] (\ax,\ay) -- (\bx,\by);}\fi
  \foreach \n/\px/\py in {#3}{\node[circle, fill=black!75, draw=white, line width=0.6pt, text=white, inner sep=0pt, minimum size=0.32cm, font=\sffamily\tiny\bfseries] at (\px,\py) {\n};}
  \end{scope}\end{tikzpicture}}
\begin{tabular}{@{}c@{\hspace{5pt}}c@{\hspace{5pt}}c@{\hspace{5pt}}c@{}}
\uipanel{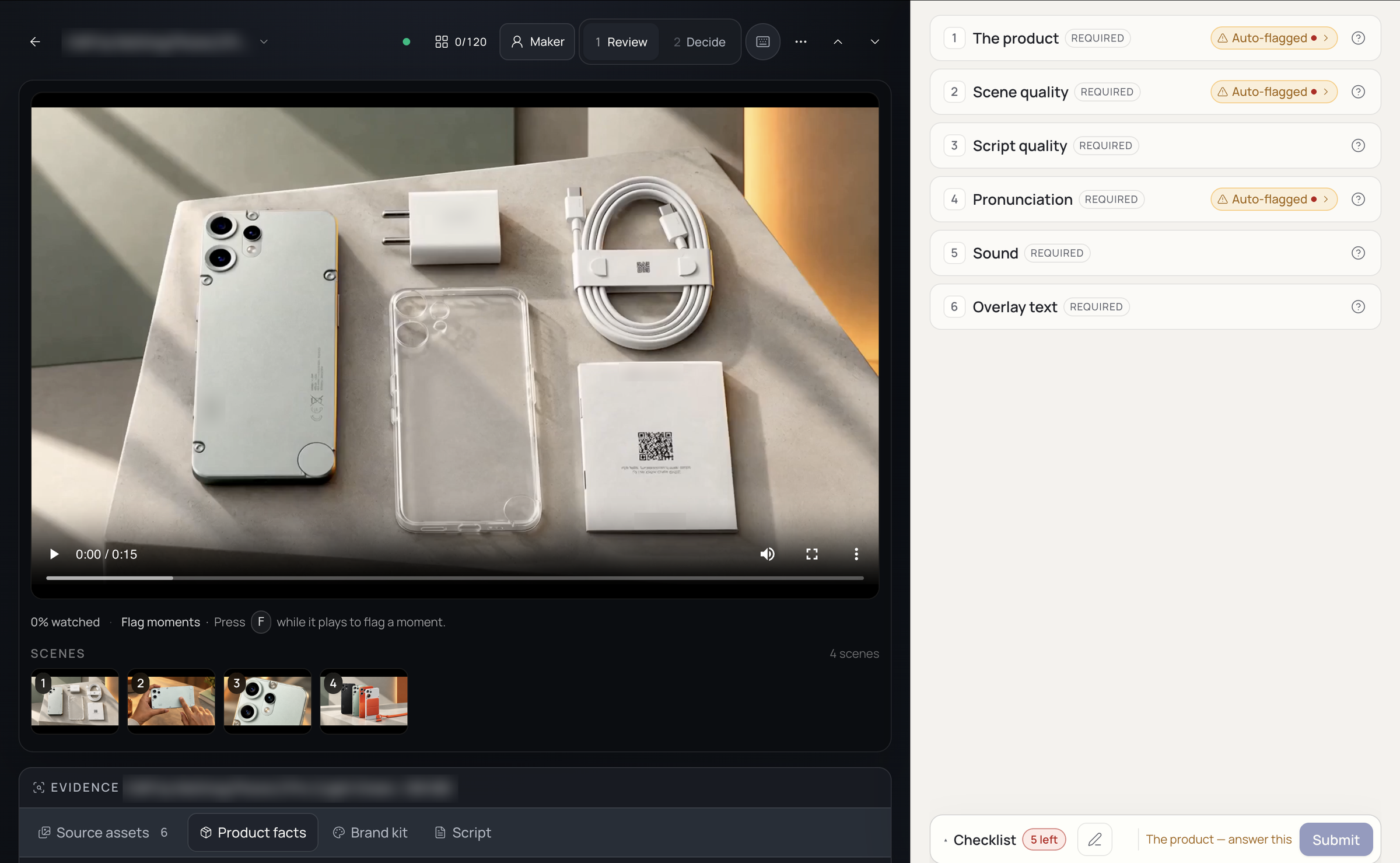}{1/0.30/0.19, 2/0.66/0.83, 3/0.375/0.075} &
\uipanel{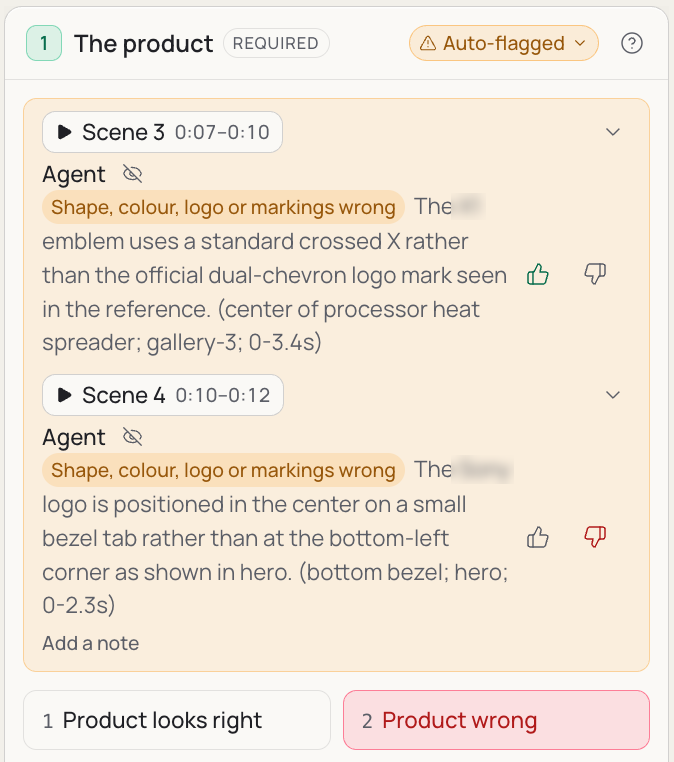}{4/0.40/0.772, 5/0.50/0.055} &
\uipanel{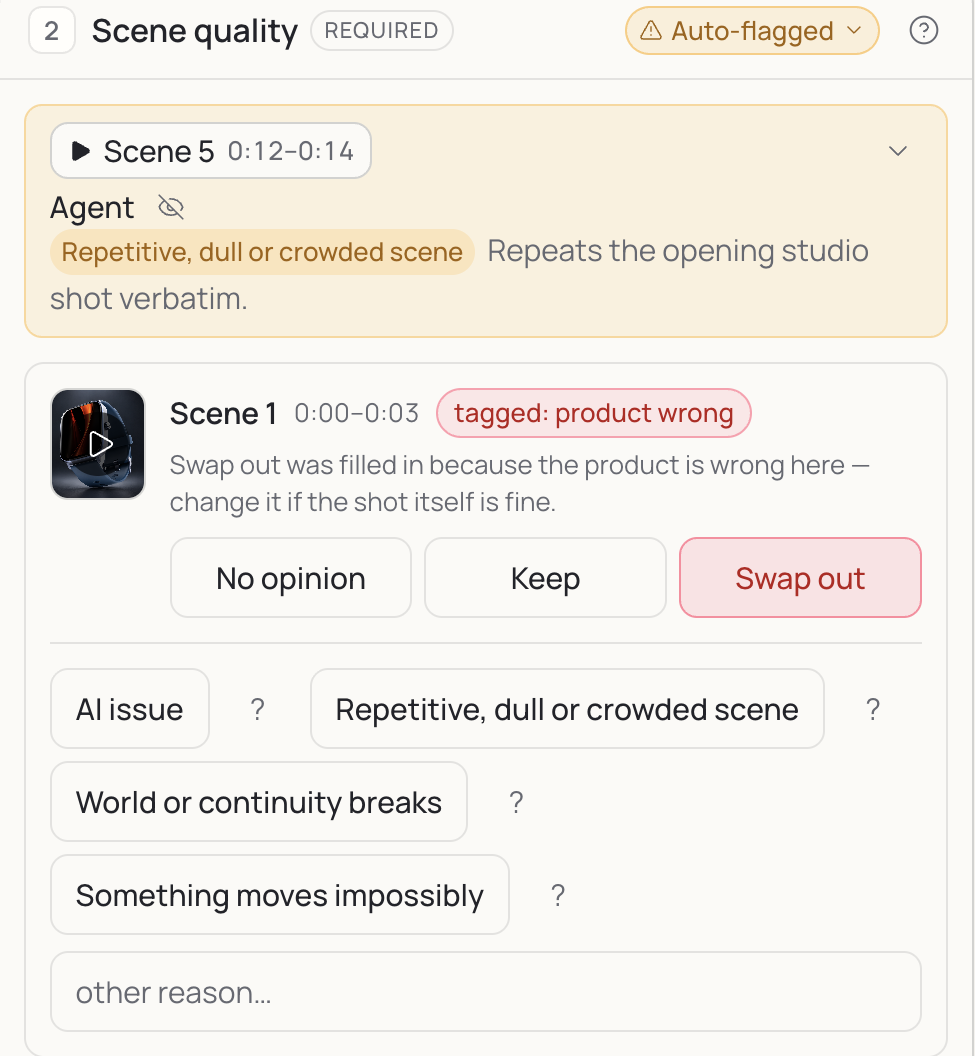}{6/0.451/0.820, 7/0.10/0.453, 8/0.656/0.197} &
\uipanel{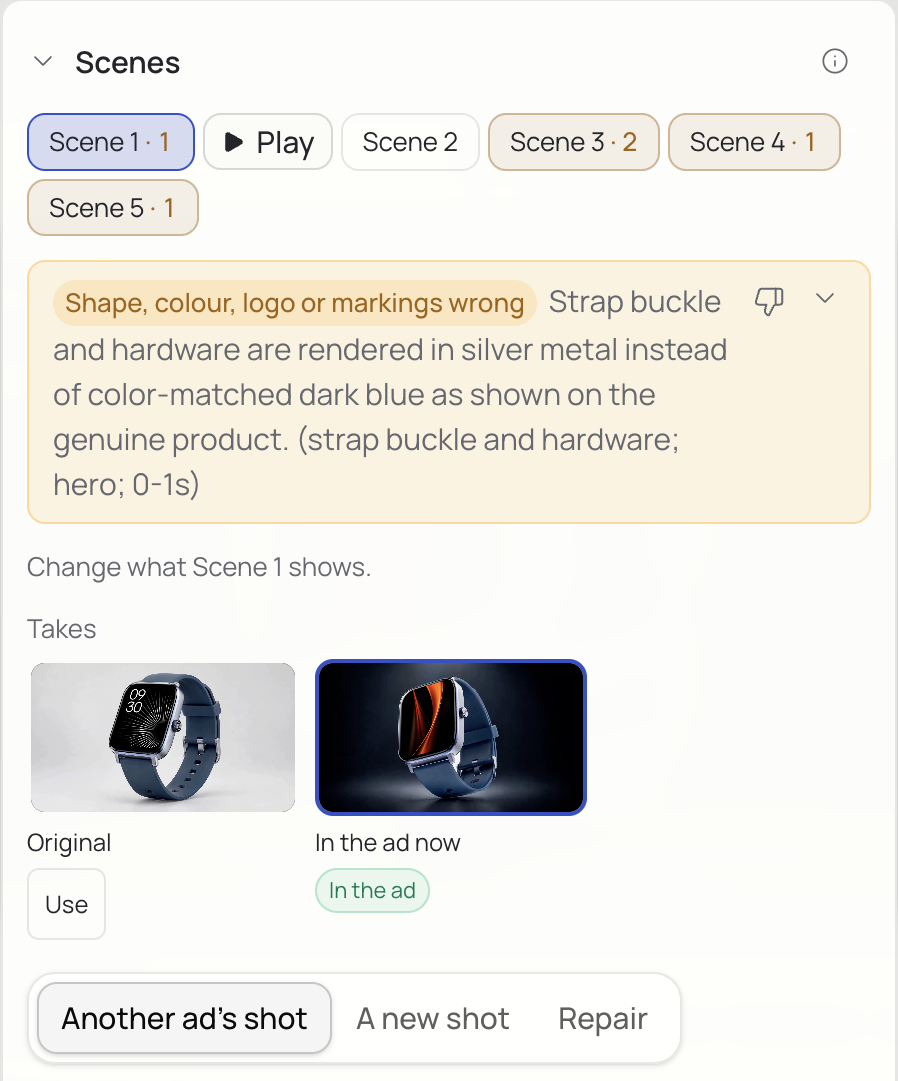}{9/0.30/0.808, 10/0.88/0.60, 11/0.75/0.32, 12/0.84/0.058} \\
\footnotesize (a) Review page & \footnotesize (b) Product axis & \footnotesize (c) Scene quality axis & \footnotesize (d) Edit page
\end{tabular}
\caption{The production portal. Panels (a) to (c) show the review and panel (d) the edit page. (a) The review page for one ad: \cnum{1} the ad and its scenes, \cnum{2} the six questions, collapsed, each tagged when a judge flagged it, and \cnum{3} the evidence: the catalog photos, product facts, brand kit and script. (b) \cnum{4} A judge's finding sits inside the question with its scene, kind and evidence; the reviewer accepts or rejects each with a thumbs up or down and \cnum{5} answers below, where ``Product wrong'' then asks which scenes are at fault. (c) \cnum{6} The scene judges' finding above \cnum{7} the per-scene grid, with keep, swap out or no opinion for each scene, and \cnum{8} the secondary options for the kind of problem. (d) On the edit page: \cnum{9} the scenes with confirmed issues; \cnum{10} a confirmed issue opens the scene tool on its scene, with \cnum{11} the scene's versions and \cnum{12} ways to change it: another ad's scene of the same product, a new scene, or Repair, which fixes the scene in place, today by regenerating its clip at higher resolution.}
\label{fig:portal}
\end{figure*}

\paragraph{What a reviewer sees with and without judges.}
Without judges, a reviewer sees each question, the catalog photos, the product's listing facts and the scene, line and caption pickers. With judges (\cref{fig:portal}), a flagged question also shows an ``Auto-flagged'' tag and the findings for that axis, grouped by scene and signed ``Agent''. A thumbs up writes the finding into the answer; a thumbs down records a rejection with an optional note. Reviewers see only their own thumbs. A judge's flag makes its question required on that ad, and a review can be submitted only once nearly all of the ad has played, every required question is answered and every finding has a thumb.

\paragraph{From review to decision.}
Rules turn each review into one of three verdicts: approve, edit or reject. Each answer, and most secondary options, maps to a problem category with a severity; pronunciation and sound share one category. A product marked wrong in any scene rejects the ad, and so do issues in two categories or two scenes to replace; an issue in one category or one scene to replace sends the ad to editing (\cref{sec:fixing}), and an ad with no issues is approved. These thresholds are policy choices; a person can overrule the verdict, for example sending an ad with a product fault for editing. By the majority's marks, the product rule alone would reject 117 of the round's 120 ads. The queue puts ads with the fewest, lightest issues first, so a few usable ads per product are ready quickly.

\section{Experiments: AI Judges and Reviewers}
\label{sec:portal}

\paragraph{Setup.}
We ran one review round on 120 generated 15-second ads from our pipeline (562 scenes, 20 products). Five internal reviewers used the portal with the judges' findings (\cref{sec:judges}) on screen and every question required. Of the 120 ads, 78 had three reviews, 2 had two and 40 had one; a majority means more than half of an ad's reviewers, so on a one-review ad it is that reviewer. Two reviewers then reviewed 60 of these ads with every judge hidden: R1, new to the ads, and R2, a round reviewer. To score the judges, we also use blind labels, made by labelers who had not seen the judge's output (\cref{tab:labelsets}): a judge's recall is the share of defective items it flags, and its precision the share of its flags that are defective.

\begin{figure}[t]
\centering
\includegraphics[width=\columnwidth]{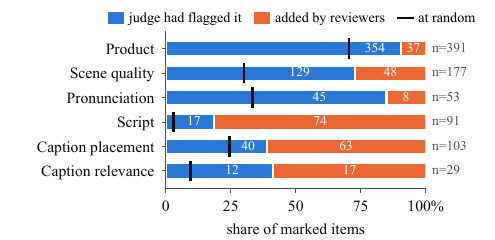}
\caption{Each bar splits the items (scenes, lines or captions) any reviewer marked on its axis into those a judge had flagged (blue) and those reviewers added (orange), in the 120-ad round. Black ticks: the share random flags at the judge's flag rate would cover.}
\label{fig:split}
\end{figure}

\paragraph{Judges and reviewers together.}
In the round, the majority marked the product wrong on 60\% of the 562 scenes, a script problem on 6\% and caption placement and relevance on 7\% and 2\% of 530 captions; \cref{fig:split} counts marks by any reviewer. Of the scenes any reviewer marked product wrong, the product judge had flagged 91\% (blue in \cref{fig:split}). It flagged 70\% of all scenes, so flagging that many at random would already catch 70\% of the marked scenes (black tick). The scene-quality and pronunciation judges had also flagged most marked items. A thumbs up writes a finding into the mark, so these shares show where marks came from and say little about a judge's accuracy (\cref{tab:validity}); reviewers rejected 153 of the pronunciation judge's 187 flags. On the script, reviewers found most problems themselves (orange).

\begin{figure}[t]
\centering
\includegraphics[width=\columnwidth]{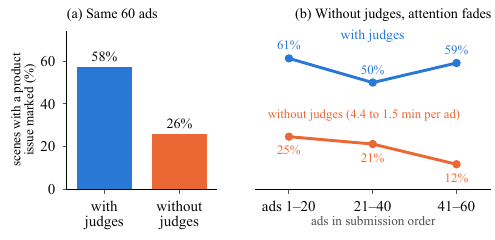}
\caption{Product marks with judge findings on screen and with judges hidden, same 60 ads. (a) Share of scenes marked product wrong: the round's majority with judges, the reviewers' average without. (b) R2, who saw these ads both ways.}
\label{fig:ww}
\end{figure}

\paragraph{The same ads with judges hidden.}
On 60 of the round's ads (280 scenes), reviewers with judges hidden marked the product wrong on 26\% of scenes on average, against 58\% for the round's majority with judge findings on screen (\cref{fig:ww}a; gap 31 points, 95\% interval [24, 39]; pre-registered for each reviewer, averaged post hoc; \cref{app:alone}). On the 39 of these ads with three reviews in the round, the gap held (61\% against 29\%; post hoc). R2 had reviewed the same ads in the round two weeks earlier and, with judge findings on screen, had marked the product wrong on 57\% of the scenes, against their own 19\% with judges hidden; memory of the findings would, if anything, narrow that gap. Only the product shows this gap (other axes: \cref{tab:alone}). Of the scenes the majority marked, R1, who had not seen the ads, marked 38\% and the product judge flagged 96\%.

\paragraph{Without judges, one reviewer's attention faded.}
In an exploratory analysis, with judges hidden, R2, who saw the ads both ways, marked the product wrong on 25\%~$\to$~21\%~$\to$~12\% of scenes over their first, middle and last 20 ads, while the median review fell from 4.4 to 1.9 to 1.5 minutes; with judge findings, R2 marked the same ads at a steady 61\%~$\to$~50\%~$\to$~59\% (\cref{fig:ww}b). The findings also brought reviewers level: with judges hidden, R2 marked far fewer scenes than R1, but with judge findings R2 marked 59\% of scenes on ads other round reviewers also reviewed, close to those reviewers' 64\%.

\paragraph{Which findings reviewers accept.}
How often reviewers accept a finding varies with the kind of problem it names. They accepted the product judge's garbled-text findings in 90\% of their votes and its findings on shape, color, logo or markings in 61\%, and each of the five reviewers accepted garbled-text findings more often. Accepting a finding copies its kind onto the reviewer's mark, and reviewers kept that kind on 91\% of such marks, so the kinds on product marks come mostly from the judge. Overall, reviewers accepted 75\% of product findings, while the judge's precision on blind labels was 0.46. The two need not conflict: a judge's flags are right more often when more scenes are defective, and the round likely had more defective scenes, since the judge flagged 70\% of the round's scenes against 47\% on blind labels. As the round has no blind labels, we cannot tell how many added marks are right or how much acceptance is over-trust.

\paragraph{What a review costs.}
On the same 60 ads, a review took a median of 4.1 minutes with judge findings and 2.9 with judges hidden (\cref{app:alone}), about \$1.40 against \$1.00 per review at \$20 an hour. In an exploratory check, more time did not close the product gap: reviews with judges hidden that took at least 4.1 minutes still marked only 35\% of scenes, against 65\% for the majority on the same ads.

\paragraph{Can a judge decide alone?}
Not yet. Product defects are subtle and vary by product, and a missed one can reach customers; a false flag costs one click to reject. So we tune the product judge for recall: it catches nearly all defects, and on blind labels about half its flags were right (\cref{tab:validity}), so reviewers check each one. A precision-heavy setting would be right on 71\% of its flags but catch only 15 of 29 defective scenes (\cref{app:baselines}). Judges also miss problems: skipping the script question wherever its judge passed would have dropped it from 60 of 88 reviews where reviewers marked a script problem, so the portal asks every question.

\paragraph{Where reviewers agree.}
On ads with three reviews and judge findings on screen, reviewers agreed well on the product and on which scenes to replace (Fleiss' $\kappa$ 0.61 and 0.80; 1 is full agreement, 0 is chance) and poorly on the script and caption placement (0.26 and 0.11). Seeing the same judge findings may raise this agreement. Unflagged product problems seem easy to miss: on scenes more than one reviewer saw, 90\% of the product problems the judge had not flagged were marked by a single reviewer (26 of 29), against 12\% of flagged ones (30 of 260).

\begin{table}[t]
\renewcommand{\arraystretch}{0.9}
\centering\footnotesize
\caption{Evidence on each judge. Blind labels were made without seeing the judge's output (\cref{tab:labelsets}). Recall: share of defective items the judge flagged; precision: share of its flags that were defective. Accepted: share of the judge's findings in the round that most reviewers accepted; dashes: no comparable figure (\cref{tab:labelsets}).}
\label{tab:validity}
\setlength{\tabcolsep}{2.4pt}\renewcommand{\arraystretch}{1.1}
\begin{tabular}{@{}l r r r r@{}}
\toprule
& \multicolumn{3}{c}{\hd{Blind labels}} & \hd{Round} \\
\cmidrule(lr){2-4}\cmidrule(l){5-5}
\hd{Judge} & \hd{Items (defective)} & \hd{Recall} & \hd{Precision} & \hd{Accepted} \\
\midrule
Product & 120 scenes (29) & 26 of 29 & 26 of 56 & 75\% \\
\zebra Pronunciation & 208 lines (43) & 28 of 43 & 28 of 90 & -- \\
Caption placement & 530 captions (39) & 15 of 39 & 15 of 130 & -- \\
\zebra Caption relevance & 530 captions (11) & 4 of 11 & 4 of 51 & -- \\
Scene quality & \multicolumn{3}{c}{no blind labels} & 96\% \\
\zebra Script & \multicolumn{3}{c}{no blind labels} & 94\% \\
\bottomrule
\end{tabular}

\end{table}

\section{From Finding to Fix}
\label{sec:fixing}

Once reviewers confirm an issue, the fix should correct only what is wrong and leave the rest of the ad alone. Each issue names its axis, the kind of problem and the unit at fault: a scene, a line, a caption or the whole ad. Three rules keep each fix local and checkable (\cref{tab:fixes}), and \cref{app:gallery} shows fixes before and after.

\begin{table*}[t]
\centering\footnotesize
\newcommand{\how}[1]{\par{\scriptsize\color{black!72}\baselineskip=8.5pt #1\par}}
\newcommand{\grp}{\arrayrulecolor{black!30}\specialrule{0.25pt}{0.3pt}{0.8pt}\arrayrulecolor{black}}
\newcommand{\sub}{\addlinespace[2pt]}
\newcolumntype{P}[1]{>{\raggedright\arraybackslash\setlength{\baselineskip}{8.5pt}}p{#1}}
\caption{The fix for each problem and the kind of model it uses, with representative models cited. At generation, the script fix also runs automatically on the lines the script judge quotes; \cref{app:models} describes the kinds of models each fix can use, and \cref{app:gallery} shows examples before and after.}
\label{tab:fixes}
\setlength{\tabcolsep}{4pt}
\setlength{\arrayrulewidth}{0.25pt}
\renewcommand{\arraystretch}{1.08}
\begin{tabularx}{\textwidth}{@{}>{\bfseries}l >{\raggedright\arraybackslash}p{4.3cm} >{\raggedright\arraybackslash}X >{\raggedright\arraybackslash}p{5.1cm}@{}}
\toprule
\hd{Axis} & \hd{Problem} & \hd{Fix} & \hd{Model} \\
\midrule
Product & garbled text & Repair the scene in place at higher resolution & generative upscaler (video-to-video)~\cite{wang2026seedvr2,xie2025star,topazlabs2025starlight} \\
\grp
\multirow[t]{3}{*}{\begin{tabular}[t]{@{}l@{}}Product,\\scene quality\end{tabular}} & \multirow[t]{3}{4.3cm}{\raggedright any other product or scene-quality problem (\cref{tab:axes})} & Borrow an unflagged scene from another ad & none; ranked by fit, judge-filtered \\
 & & Generate a new scene from the same references & reference-to-video~\cite{seedance2026seedance2,jiang2025vace,liu2025phantom} \\
 & & Edit the clip & video-to-video editing~\cite{google2026geminiomni,klingteam2025klingomni,jiang2025vace} \\
\grp
Pronunciation & mispronounced word & Respell the word and re-voice only its line & text-to-speech~\cite{meshram2026gemini31flashtts,elevenlabs2025v3,du2024cosyvoice2} \\
\grp
Script & unsupported claim & Correct the claim's words to match the listing, or regenerate the line & language model; text-to-speech \\
 & awkward phrasing & Rephrase or regenerate the line & language model; text-to-speech \\
\grp
Sound & voice & Re-voice the ad, keeping its timing & speech-to-speech~\cite{elevenlabs2026voicechanger,liu2024seedvc,baas2023knnvc} \\
 & music & Replace the music bed from an editable prompt & text-to-music~\cite{deepmind2026lyria35,copet2023musicgen,evans2025stableaudioopen} \\
\grp
Captions & placement & Move or restyle the caption on the overlay layer on top of the video & vision-language model \\
 & relevance & Reword or remove the caption, or replace the logo & language model \\
\bottomrule
\end{tabularx}
\end{table*}

\begin{figure}[b]
\centering
{\setlength{\fboxsep}{0pt}\textcolor{black!30}{\fbox{\includegraphics[width=0.9\columnwidth]{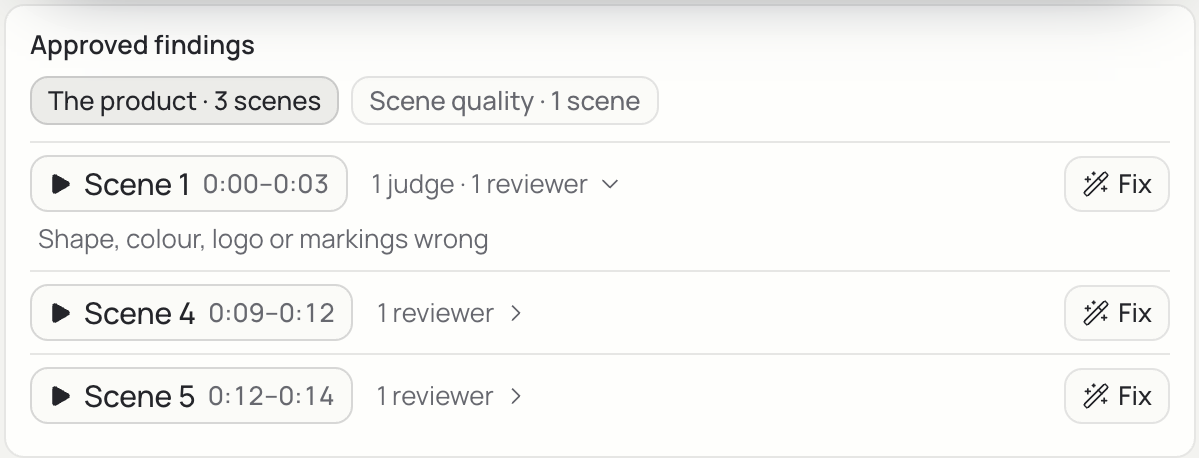}}}}
\caption{The decide board for one ad (cropped): confirmed issues by axis and scene, who backed each, and Fix, which opens the matching tool.}
\label{fig:decide}
\end{figure}

\textbf{Route to the smallest unit.}\enspace{}On the decide board (\cref{fig:decide}), which lists an ad's confirmed issues, each issue's Fix button opens the tool for its axis and kind on its own scene or line (\cref{fig:portal}d). Garbled text goes to Repair, a wrong product to a borrowed, new or edited scene, and a mispronounced word to its own line, since re-voicing the whole ad converts the old speech and would keep the mistake; a finding no reviewer accepted opens nothing.

\textbf{Leave the rest alone.}\enspace{}The ad is stored as separate clips, lines, captions, logo and music. A fix rewrites only the part at fault, at its old length, and keeps everything else as it was: a new or borrowed scene is trimmed or padded to its slot and a line is fitted to its slot, so narration and later scenes keep their timing. Untouched lines stay byte-identical, and the final render re-encodes only the changed scenes.

\textbf{Review the edited ad afresh.}\enspace{}Sending the edited ad back to review (the edit page's Publish button) renders it, the findings on the old version go stale, and the edited ad is reviewed again like a new ad.

\textbf{What the tests show.}\enspace{}On 26 rated scenes with garbled product text, Repair, whose engine regenerates the scene's own clip at higher resolution, guided by its original prompt and references, fixed the text fully in 18 and partly in 8, and kept the same scene in all 26. The script fix also ran automatically at generation: on 410 earlier ads, after only the lines the script judge quoted were rewritten (`a full year of battery life', for example, became the listing's `up to 12 months'), the judge's own re-read cleared 129 of the 134 scripts it had flagged; two annotators found 78\% of these clean, and after adjudication 7 of the 129 still had a defect the judge missed (\cref{tab:labelsets}).

\section{Judges for Each Axis}
\label{sec:judges}

Each axis has its own judge, since the axes need different evidence (\cref{tab:axes}). Each gets the evidence a careful reviewer would use and points its findings at a scene, line or quoted span; \cref{tab:experiments} lists the experiments that changed a decision, and \cref{app:models} names the closed models the judges run on.

\begin{figure}[b]
\centering
\includegraphics{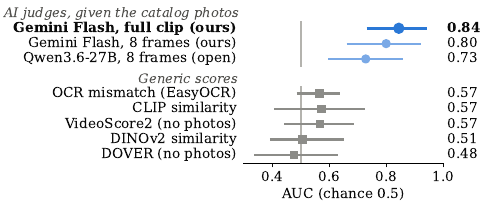}
\caption{AI judges given the catalog photos against generic scores on 120 scenes labeled blind. Only the AI judges' intervals lie entirely above chance (0.5). Lines: 95\% bootstrap intervals (paired gaps: \cref{tab:baselines}).}
\label{fig:auc}
\end{figure}

\textbf{Product.}\enspace{}The judge watches each clip with every catalog photo (\cref{tab:axes}). On 120 scenes labeled blind from the clip, its 0 to 100 fidelity score separated defective from clean scenes (AUC 0.84), while generic similarity, OCR and video-quality~\cite{wu2023dover,he2025videoscore2} scores stayed near chance (\cref{fig:auc}). On eight frames the same judge reached 0.80 (paired gap [$-$0.05, 0.15]), so we cannot say the motion helps; the best open VLM~\cite{qwen2026qwen35,qwen2026qwen36} with our prompt and photos reached 0.73, not significantly below ours. The judge's design was chosen on these labels and no baseline was tuned, so its lead is optimistic; even so, the 95\% interval of every paired gap to the generic scores excludes zero (\cref{tab:baselines}). Four annotator accounts made the labels before any judge existed (\cref{tab:labelsets}; pairwise $\kappa$ about 0.51), excluding 26 split-vote scenes, likely hard cases.

\begin{table*}[t]
\centering\footnotesize
\newcommand{\mk}[1]{\textsf{\textbf{#1}}}
\caption{Experiments that shaped the judges: each row asks one design question and gives the answer and its evidence. Labels: \mk{B} labelers blind to judge output, \mk{A} reviewers who saw it, \mk{M} a model or measurement; numbers on \mk{A} and \mk{M} labels measure agreement with that reference only. Choices were made and scored on the same labels, so numbers are optimistic; brackets are 95\% intervals.}
\label{tab:experiments}
\setlength{\tabcolsep}{3.5pt}
\renewcommand{\arraystretch}{1.15}
\begin{tabularx}{\textwidth}{>{\bfseries}l >{\raggedright\arraybackslash}p{5.6cm} >{\sffamily\bfseries}c@{\hspace{3pt}} >{\raggedright\arraybackslash}X}
\toprule
\hd{Axis} & \hd{Question and answer} & \multicolumn{2}{l}{\hd{Evidence}} \\
\midrule
Product & Judge the clip or its starting frame? \textbf{The clip:} fewer false flags (prompts differ). & B & 120 scenes: both catch most defects (26 and 24 of 29), but the clip judge flags fewer clean scenes (30 against 53 of 91); precision 0.46 against 0.31; $\kappa$~0.43~[0.27,~0.60] against 0.16~[0.04,~0.30], paired gap [0.10,~0.43] \\
 & Do generic scores work? \textbf{No:} near chance. & B & AUC: generic scores 0.48 to 0.57, open VLMs at most 0.73, ours 0.84 (\cref{fig:auc}) \\
\zebra Scene quality & One judge or two? \textbf{Two:} the pair finds more. & A & 489 reviewer-scene pairs: either judge flags 69\% of the scenes reviewers wanted swapped, the better one alone 55\%, at precision 0.84 against 0.89 \\
Script & Read text on listing photos? \textbf{Yes:} factual calls improve. & A & 200 ratings on 155 ads, raters saw both judge versions' calls: the call is right in 94\% of cases when the photos are read, 50\% when not \\
\zebra Captions & Can a model judge captions? \textbf{Not reliably:} its flags are stable across runs but barely match reviewers, so we measure placement. & B & 530 captions, against the reviewers' majority: $\kappa$ 0.07 for placement (between reviewers 0.11) and 0.10 for relevance; self-agreement 0.70 and 0.80 across runs. A best-of-five placement pick (\mk{M}) matched itself on 55 of 100 captions \\
\bottomrule
\end{tabularx}
\end{table*}

\textbf{Scene quality.}\enspace{}The scene judges mostly flag repeats (57 of 66 accepted findings in an earlier round), rarely exact copies: a product's catalog photos are often alike, and a studio set shown several ways looks nearly the same.

\textbf{Pronunciation.}\enspace{}The speech-to-text model must keep mispronunciations, yet many recent ones write down the word the script expects, hiding the mistake we look for: two newer models echoed the script on 30\%~\cite{google2026gemini35transcribe} and 46\%~\cite{elevenlabs2026scribev2} of defect lines, ours on 6\% (labels made after output was shown; \cref{tab:moreexperiments}). Letting the text model that rechecks each mismatch also hear the audio lowered recall, likely because it then took mispronunciations for the narrator's accent. On 208 lines labeled from audio alone, with an earlier narration engine, the judge reached recall 0.65 at precision 0.31 (\cref{tab:validity}). In the round it flagged a third of all lines (\cref{fig:split}).

\textbf{Script.}\enspace{}The script judge checks each claim against the listing, the facts stored at creation and listing-photo text, which a separate call reads first. Without the photos, the judge flagged numbers printed on packaging as unsupported claims.{} At generation a language model rewrites the quoted lines, at most twice; this ran on 41 of the round's 120 ads before narration, so their script findings follow it.

\textbf{Sound.}\enspace{}A whole-mix audio judge flagged none of 120 ads and reviewers marked sound on 2 of 278 reviews, so sound is now an optional question, shown with pronunciation. Sound problems lie in the music or the whole voice: poor recording quality, unnatural intonation, frequent mispronunciation or an accent unsuited to the market; in our experience they matter more with weaker narration engines. A blind listening study screened an earlier engine's voices for these voice problems and kept 65 of 107 (\cref{tab:moreexperiments}).

\textbf{Captions.}\enspace{}A model's pick among five placements was unstable (\cref{tab:experiments}), so we measure placement by face overlap, clutter and contrast, which were stable and told methods apart (captions over a face: 26 of 109 at fixed spots, 2 when a model placed them). On 530 captions, judges run after the round beat chance on the captions any reviewer marked (placement 39\% at a 25\% flag rate, relevance 41\% at 10\%; \cref{fig:split}) but hardly matched the majority (\cref{tab:experiments}).

\section{Limitations and Social Impact}
\label{sec:limits}

Our results come from one round of 120 ads (562 scenes) for 20 products from several categories, in one market, with English narration, and the deployed judges run on closed models. More products, less common ones and other languages would test how far the results carry; pronunciation and sound, rare problems with today's English text-to-speech, may matter more in other languages. We ran the review only on our own pipeline; other generation pipelines remain to be tried, and a version of our template that also runs the judges on any set of generated ads would make that easy. The edit side is less explored than the review. Only Repair and the script fix were tested with human ratings (\cref{sec:fixing}); experiments on the full loop from finding to edited ad to second review, with end-to-end metrics such as the share of ads delivered after edits, are future work. The judges-hidden run used two reviewers; more reviewers in a crossover would separate how much the findings help reviewers find problems from how much reviewers only follow them, and would test over-trust in the judges' findings.

\textbf{Social impact.}\enspace{}Generated video adds new ways to get a product wrong. A wrong product, label or claim misleads the shopper, hurts the seller and erodes trust in generated ads; one retailer reports that accurate product data builds customer trust and cuts returns from misrepresented products~\cite{openai2026wayfair}. Our review is meant to protect that trust. Judges flag where an ad departs from the product's own data, and a person decides on every finding before the ad is used.

\section{Conclusion}
\label{sec:conclusion}

We showed how to operationalize the review and editing of AI-generated product video ads. Six questions asked scene by scene, unchanged across a catalog, give AI judges and reviewers one shared review in which every answer points to a fix. We built a judge for each question that sees what a reviewer sees; the product judge, which watches the clip with the catalog photos, beat generic video-quality and photo-similarity scores. Reviewers with judges mark more than reviewers alone: with judge findings on screen, the round's majority marked the product wrong on more than twice as many scenes as reviewers with judges hidden. The judges cannot decide alone: we tune the product judge to catch nearly all defects, so a reviewer must reject many of its findings, and judges still miss problems, so reviewers answer every question. Each confirmed issue becomes a targeted edit of the smallest unit at fault: Repair fixes a scene's garbled product text in place, and the other tools borrow, generate or edit a scene, re-voice a line or the ad, rewrite a script line, replace the music or edit the captions. The framework is flexible: its questions stayed the same as we changed our generation models, and its edit tools draw on video, speech, music and language models, or on none. To use the review, another pipeline needs scene and line timings; to use our edit tools, it also needs each ad stored as separate clips, lines and captions. We release the review-portal template and our dataset, so others can study how reviewers use judge findings and score new judges on our clips and blind labels.

\clearpage
\newcounter{maintextpages}
\setcounter{maintextpages}{\value{page}}
\addtocounter{maintextpages}{-1}
\typeout{MAIN_TEXT_PAGES=\arabic{maintextpages}}

{
    \small
    \bibliographystyle{ieeenat_fullname}
    \bibliography{main}
}
\clearpage
\typeout{SUPP_FIRST_PAGE=\thepage}
\appendix
\onecolumn
\crefalias{section}{appendix}
\setcounter{table}{0}\renewcommand{\thetable}{S\arabic{table}}
\setcounter{figure}{0}\renewcommand{\thefigure}{S\arabic{figure}}
\begin{center}{\Large\bfseries Multimodal Judges Flag, Humans Decide: Supplementary Material}\end{center}
\providecommand{\promptfield}[1]{{\fontencoding{T1}\selectfont\texttt{\{#1\}}}}

\section{What the Review Catches}
\label{app:gallery}

\Cref{tab:gallery} shows real ads with every axis the review asks about (\cref{tab:axes}) and every kind of problem but the music: frames from the 120-ad round and, where the round has no clear case, from production runs or our studies. A product defect is shown beside the product's catalog photo. The Fix column gives the fix for the problem (\cref{tab:fixes}); where the scene was actually fixed, it shows the result. Each row says who caught the problem; in the round reviewers saw the judges' findings, so a judge flag and the marks of reviewers who accepted it count as one agreement. The defects shown come from our generation pipeline and are not in the real products; no brand shown is affiliated with this work or endorses it. Faces are blurred. The icon at the end of a row opens the example on our public gallery page (\faPlayCircle{} with its clip, \faVolumeUp{} with its audio, \faImage{} with its images only).

\newcommand{\gpic}[3]{\begin{minipage}[t]{#1}\vspace{0pt}\centering\includegraphics[width=#1]{#2}\ifx\relax#3\relax\else\\[-1pt]{\fontsize{5.5}{6.5}\selectfont\color{black!60}#3}\fi\end{minipage}}
{\setlength{\tabcolsep}{3pt}\renewcommand{\arraystretch}{1.15}\footnotesize
\begin{longtable}{@{}>{\raggedright\arraybackslash}p{2.3cm} >{\raggedright\arraybackslash}p{6.3cm} >{\raggedright\arraybackslash}p{3.0cm} >{\raggedright\arraybackslash}p{5.2cm}@{}}
\caption{What the review catches: one example of each kind of problem but the music, with the fix.}\label{tab:gallery}\\
\toprule
\hd{Problem} & \hd{What the review saw} & \hd{Fix} & \hd{What is wrong, and who caught it} \\
\midrule
\endfirsthead
\toprule
\hd{Problem} & \hd{What the review saw} & \hd{Fix} & \hd{What is wrong, and who caught it} \\
\midrule
\endhead
\bottomrule
\endfoot
\multicolumn{4}{@{}l}{\textbf{Product}} \\* \addlinespace[1pt]
\textbf{Wrong product, replaced by a still} & \gpic{2.57cm}{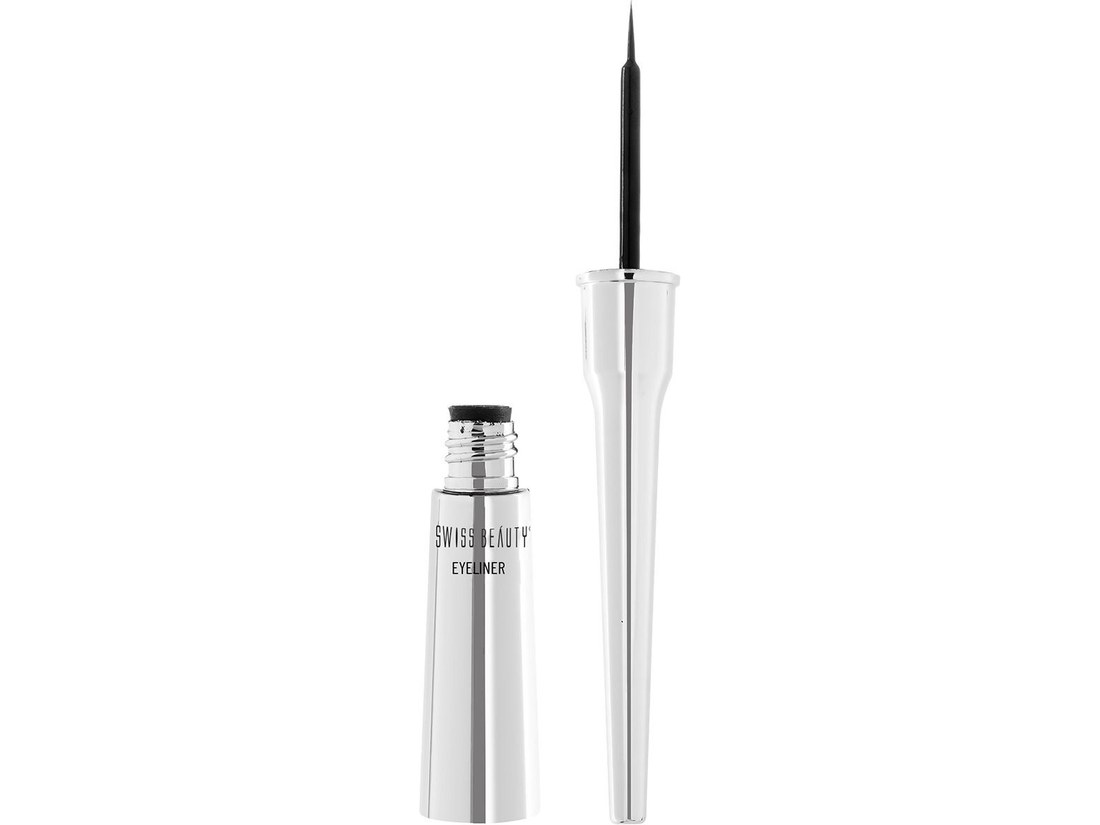}{catalog photo}\hspace{0.12cm}\gpic{3.42cm}{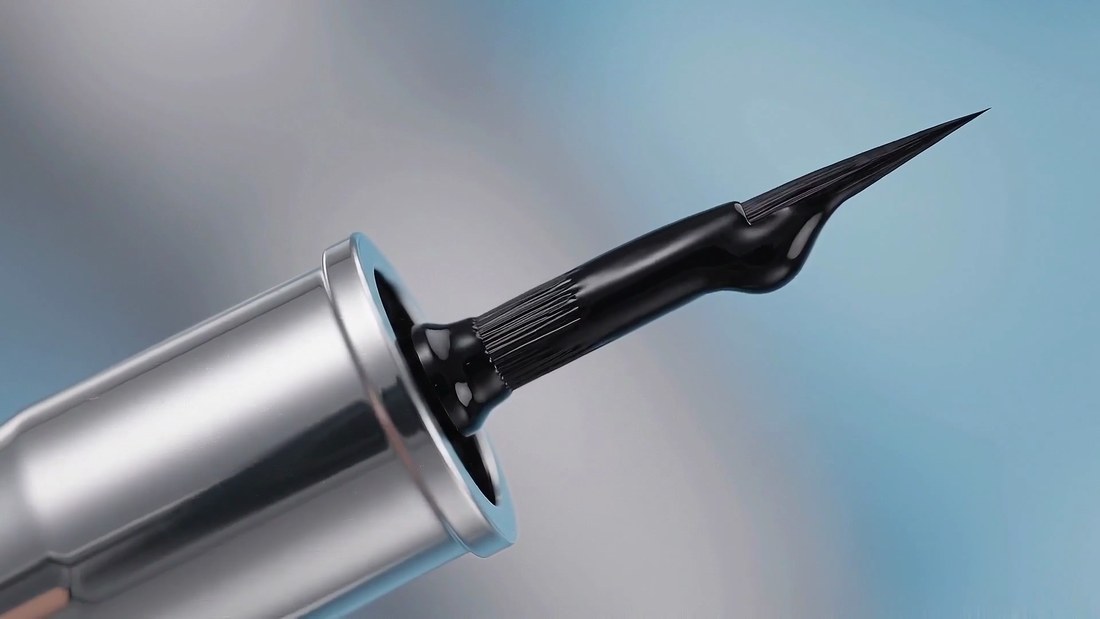}{before, 0:03} & \gpic{2.91cm}{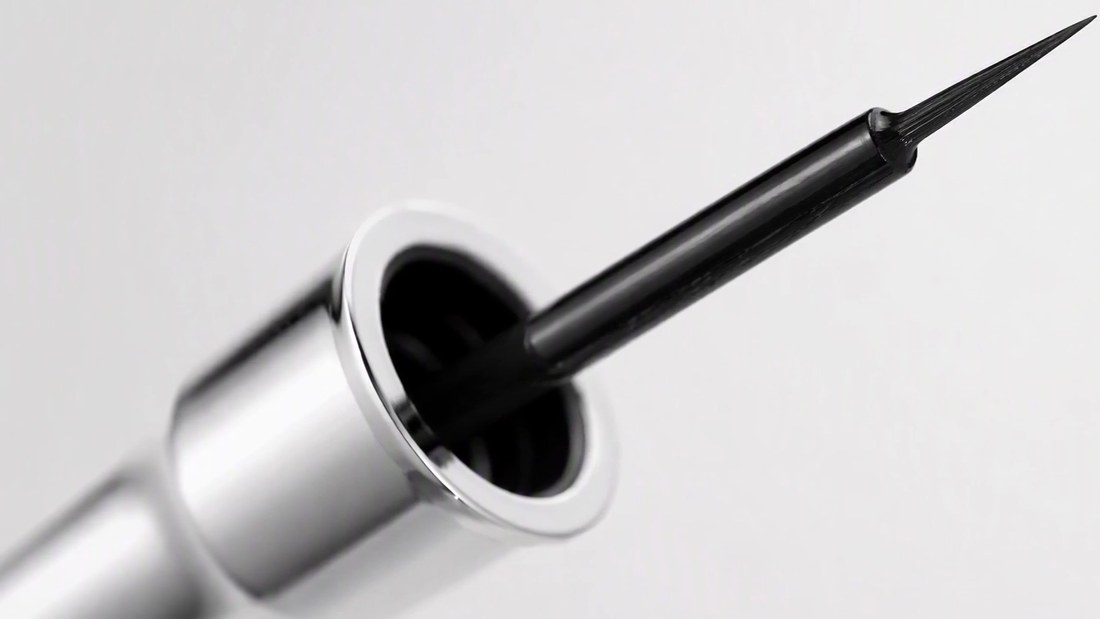}{after, 0:04}\par\vspace{1pt}{\scriptsize\textit{Replaced by an animated still}} & {\scriptsize Eyeliner ad from a production run. The applicator is drawn as a kinked, bristly brush, where the real one has a smooth precision stem (0:02-0:05); judge flagged, 1 of 1 reviewer marked. After: the scene was replaced by an animated still of the product photo (a slow pan, no generated motion), which shows a straight stem; not judged.\enspace\href{https://github.com/anoubhav/ad-review-and-edit/blob/main/gallery/README.md\#g02}{\faImage}} \\
\addlinespace[3pt]
\textbf{Shape wrong} & \gpic{2.15cm}{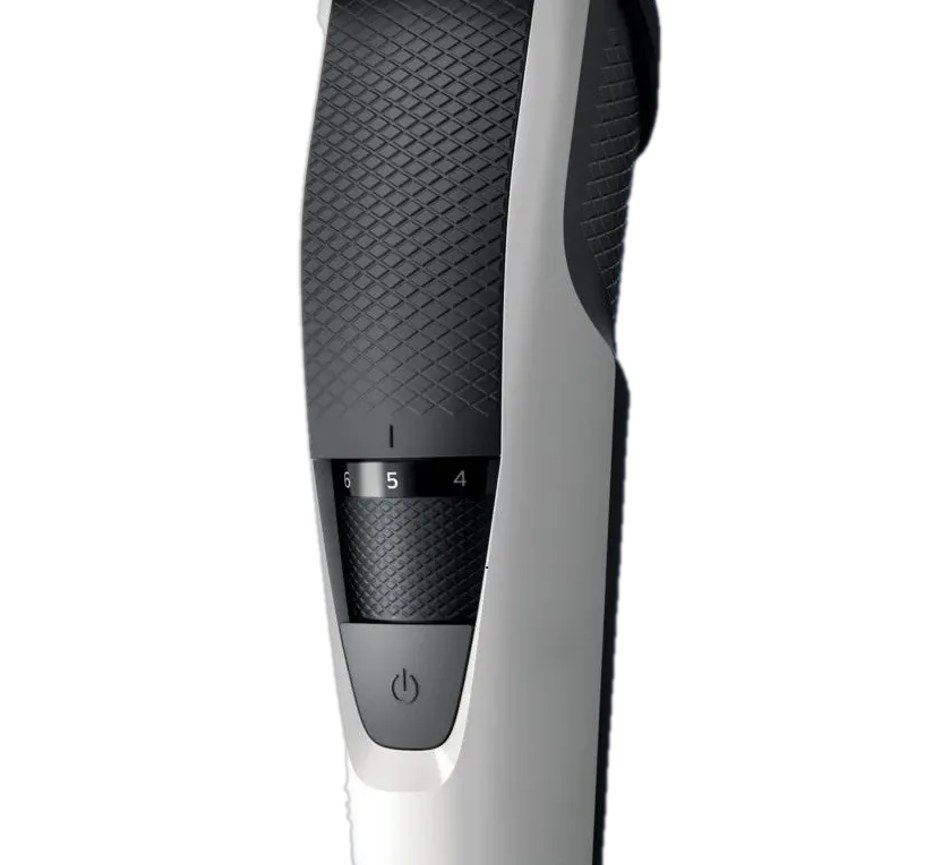}{catalog photo (crop)}\hspace{0.12cm}\gpic{3.50cm}{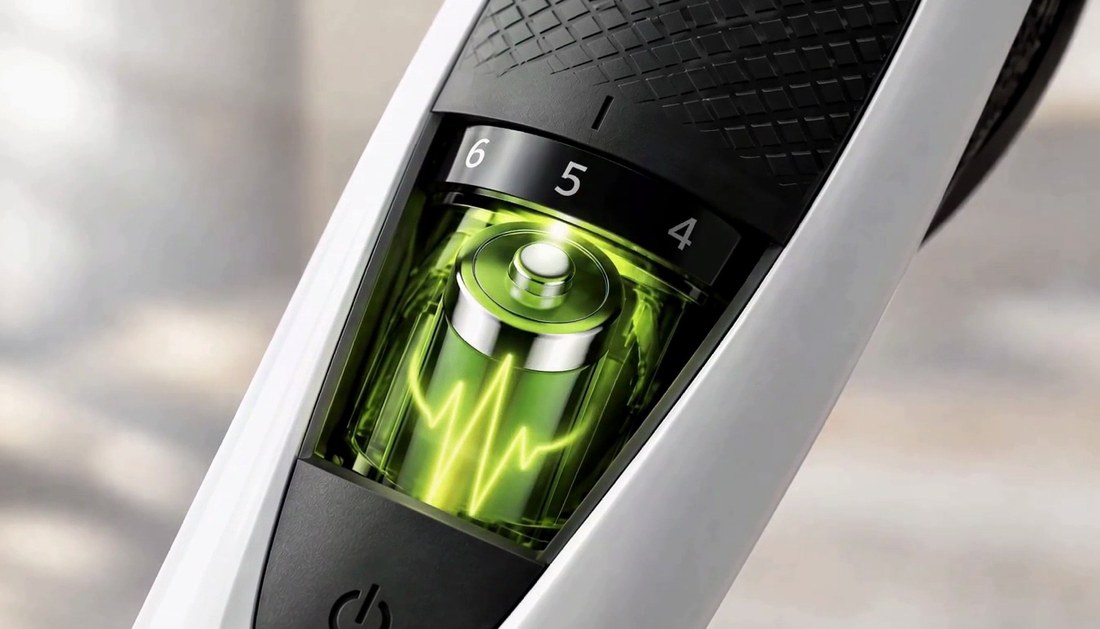}{frame 0:08} & {\scriptsize Borrow, generate or edit the scene} & {\scriptsize Beard trimmer. The ridged length-adjustment wheel is replaced by an invented see-through battery chamber (0:07-0:11). Judge flagged, 3 of 3 reviewers marked.\enspace\href{https://github.com/anoubhav/ad-review-and-edit/blob/main/gallery/README.md\#g03}{\faPlayCircle}} \\
\addlinespace[3pt]
\textbf{Color wrong} & \gpic{2.00cm}{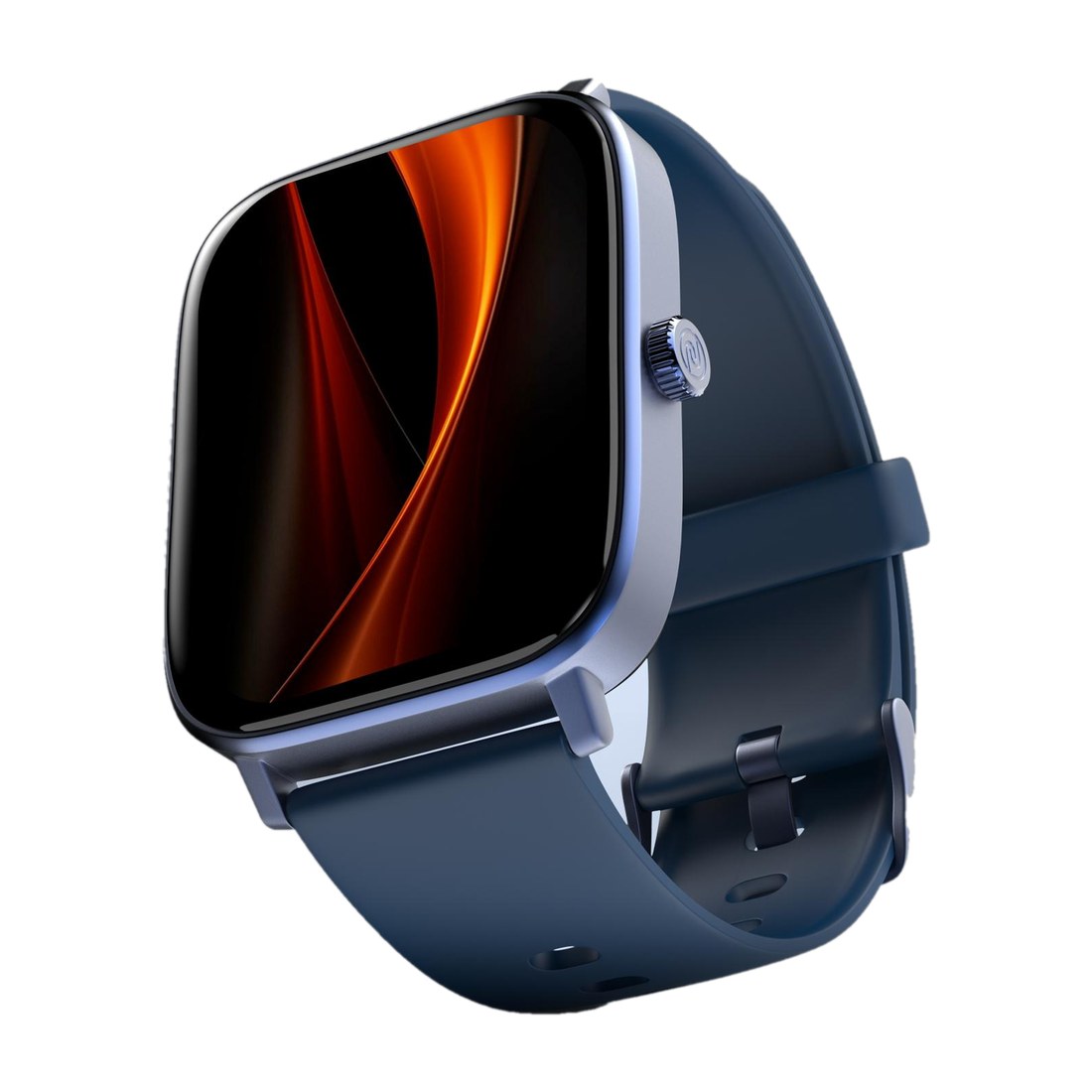}{catalog photo}\hspace{0.12cm}\gpic{3.50cm}{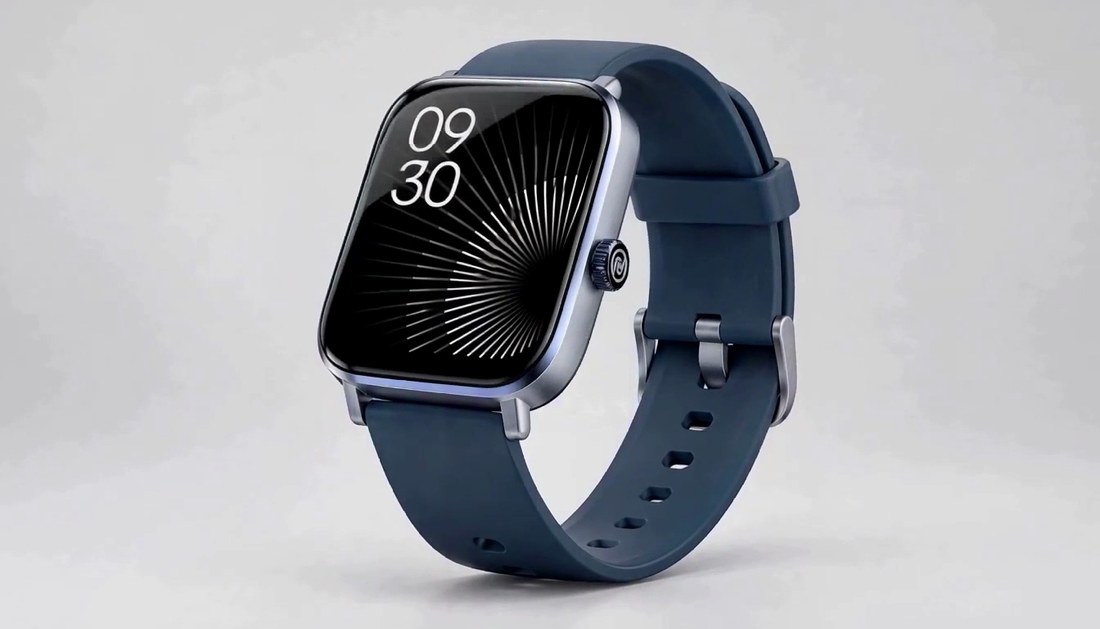}{frame 0:00} & {\scriptsize Borrow, generate or edit the scene} & {\scriptsize Smartwatch. The strap buckle is bright silver, where the real watch's is dark blue (0:00-0:01); judge flagged, 1 of 1 reviewer marked.\enspace\href{https://github.com/anoubhav/ad-review-and-edit/blob/main/gallery/README.md\#g04}{\faPlayCircle}} \\
\addlinespace[3pt]
\textbf{Text slightly garbled (low), Repair at higher resolution} & \gpic{1.87cm}{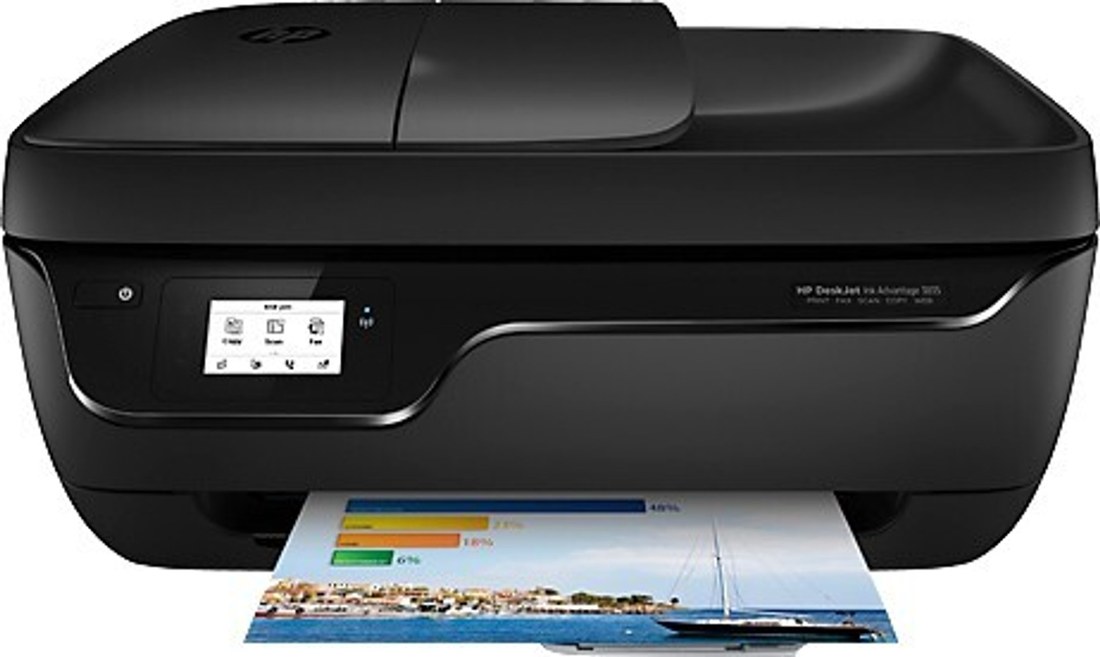}{catalog photo}\hspace{0.12cm}\gpic{4.12cm}{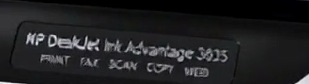}{before, 0:05 (crop)} & \gpic{2.91cm}{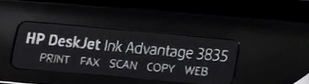}{after, Repair (crop)}\par\vspace{1pt}{\scriptsize\textit{Repair the scene in place}} & {\scriptsize Inkjet printer. Before: the model name and the small print under it, PRINT FAX SCAN COPY WEB in the catalog photo, are smeared but readable (0:05); judge flagged, 1 of 1 reviewer marked. After: Repair re-rendered the scene from its original inputs at higher resolution, and both lines read correctly; this pair is from our repair study and is judged here by eye only.\enspace\href{https://github.com/anoubhav/ad-review-and-edit/blob/main/gallery/README.md\#g06}{\faImage}} \\
\addlinespace[3pt]
\textbf{Text gibberish, color drift} & \gpic{0.86cm}{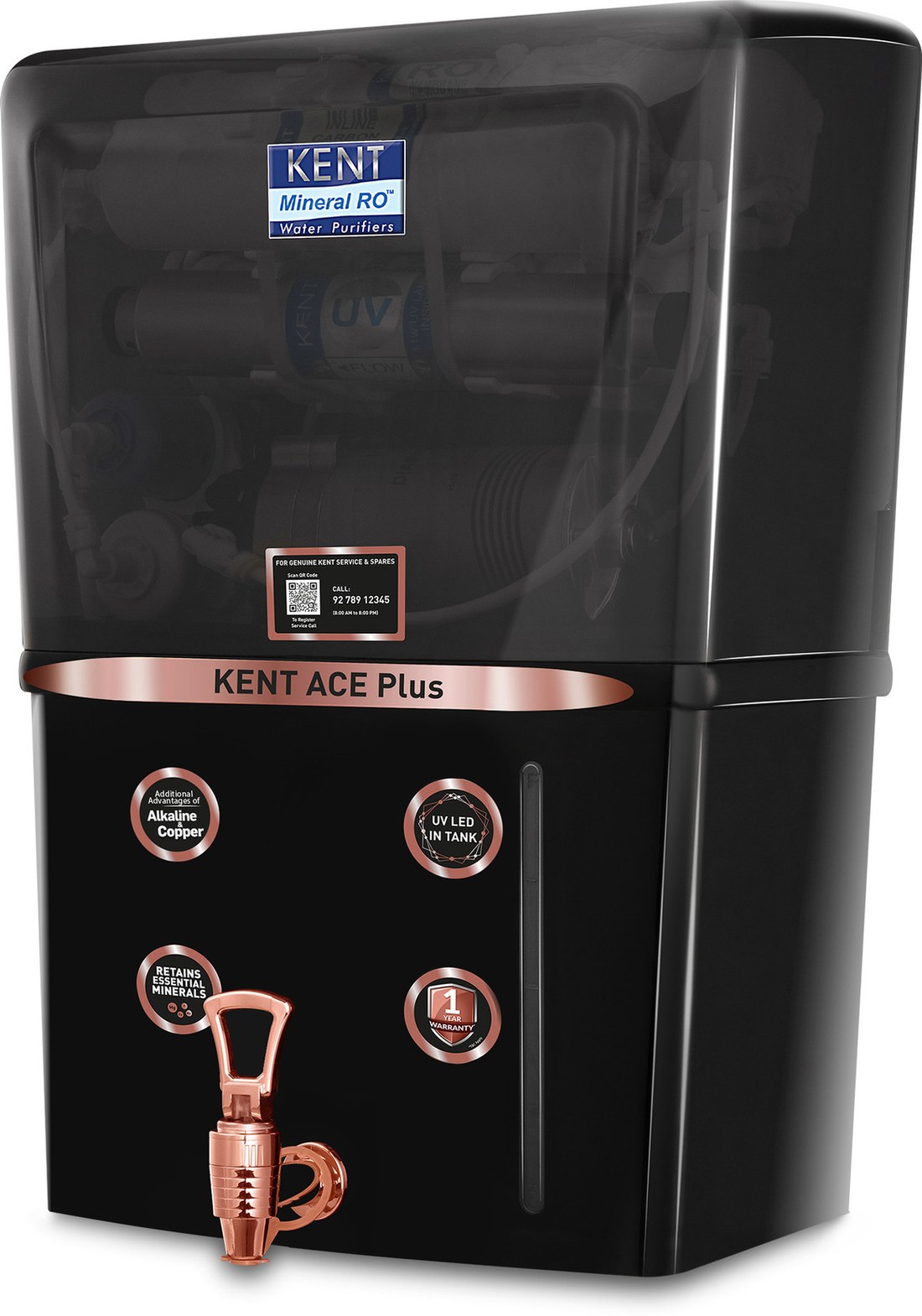}{catalog photo}\hspace{0.12cm}\gpic{2.14cm}{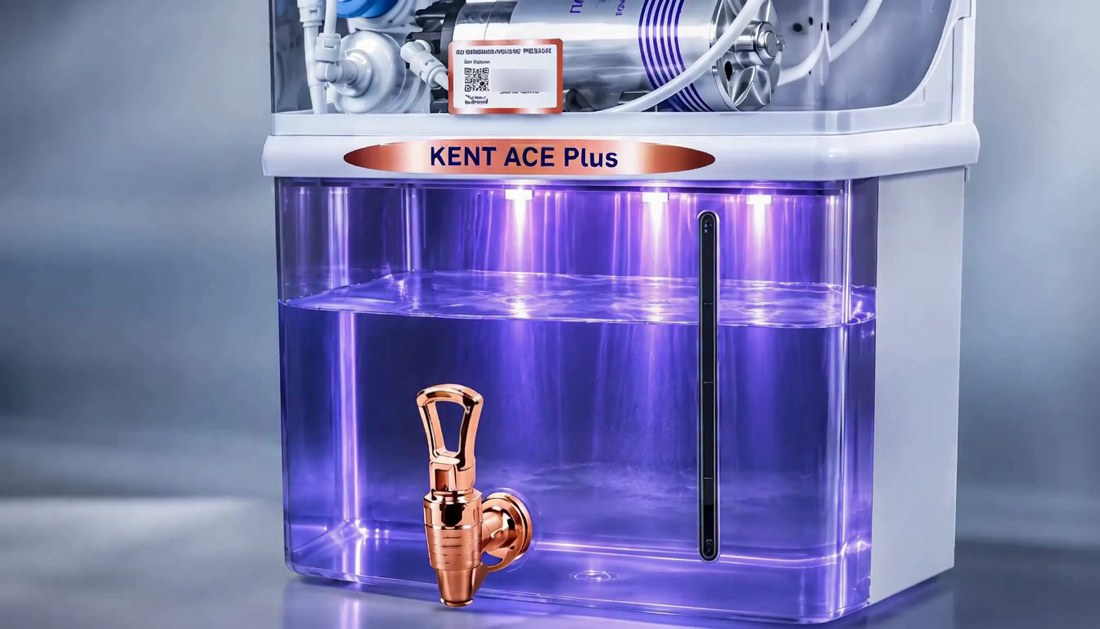}{frame 0:04 (white body; number blurred)}\hspace{0.12cm}\gpic{2.88cm}{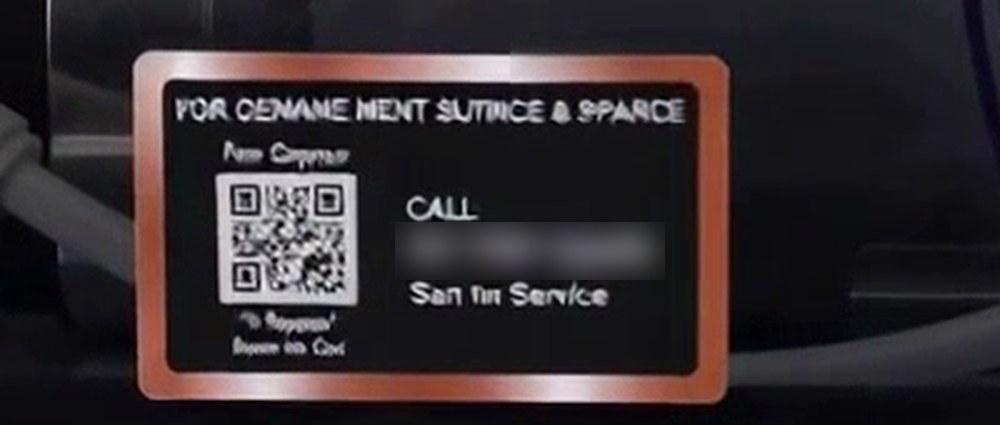}{frame 0:07, service badge (crop; number blurred)} & {\scriptsize Borrow, generate or edit the scene} & {\scriptsize Water purifier. The service badge turns into non-words with a wrong phone number, and the body starts white before turning black like the real unit (0:04-0:09). Judge flagged both, 3 of 3 reviewers marked both.\enspace\href{https://github.com/anoubhav/ad-review-and-edit/blob/main/gallery/README.md\#g07}{\faPlayCircle}} \\
\addlinespace[3pt]
\textbf{Garbled label, Repair at higher resolution} & \gpic{3.82cm}{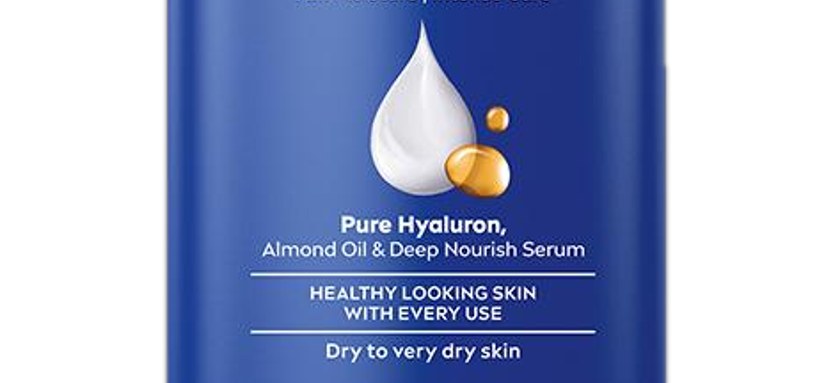}{catalog photo (crop)}\hspace{0.12cm}\gpic{2.17cm}{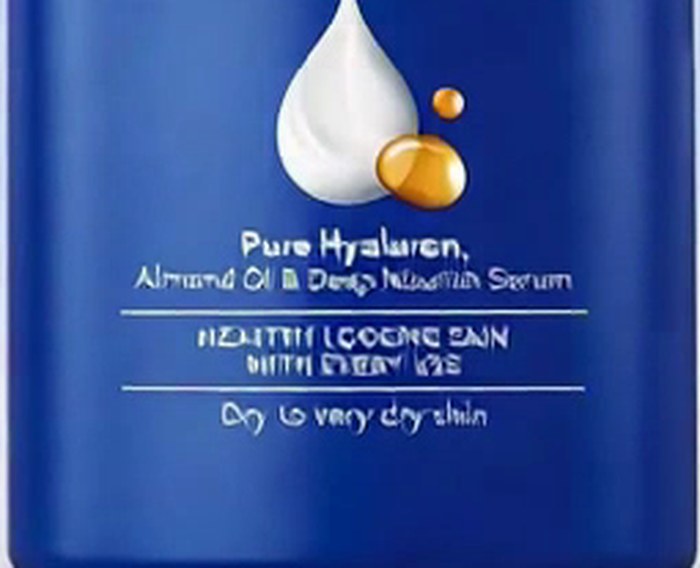}{before, 0:11 (crop)} & \gpic{2.50cm}{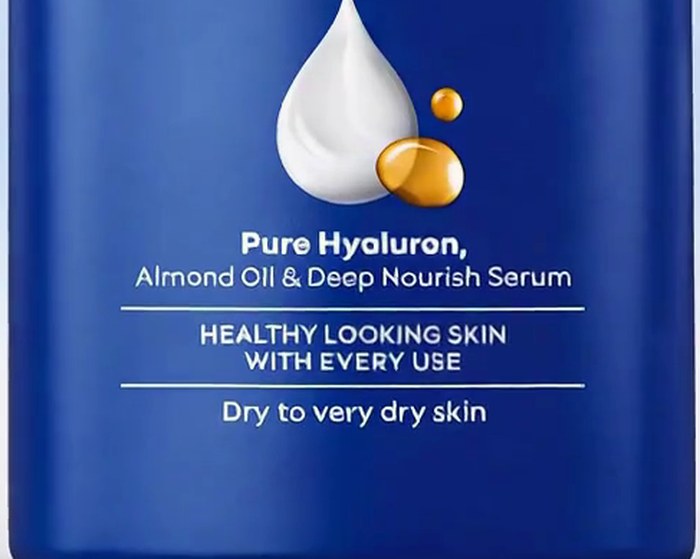}{after, Repair (crop)}\par\vspace{1pt}{\scriptsize\textit{Repair the scene in place}} & {\scriptsize Body lotion. Before: the claim lines on the lower label are smeared (0:11); judge flagged, 3 of 3 reviewers marked. After: Repair regenerated the clip itself at higher resolution, guided by its own prompt and references, and the lines read correctly; graded by four blind model graders only.\enspace\href{https://github.com/anoubhav/ad-review-and-edit/blob/main/gallery/README.md\#g08}{\faImage}} \\
\addlinespace[3pt]
\textbf{AI issue on the product} & \gpic{1.77cm}{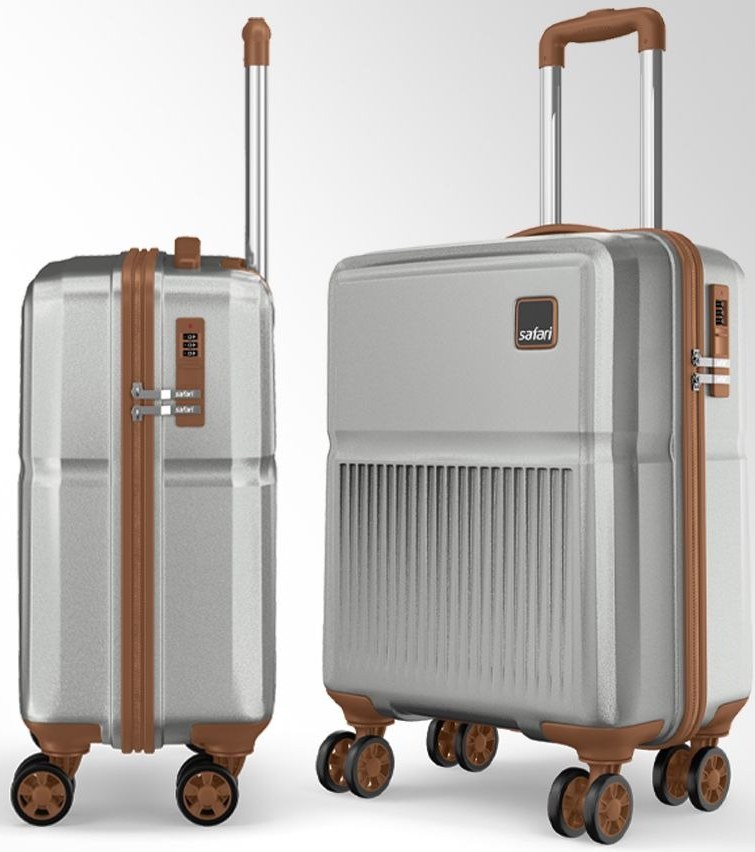}{catalog photo (crop)}\hspace{0.12cm}\gpic{3.55cm}{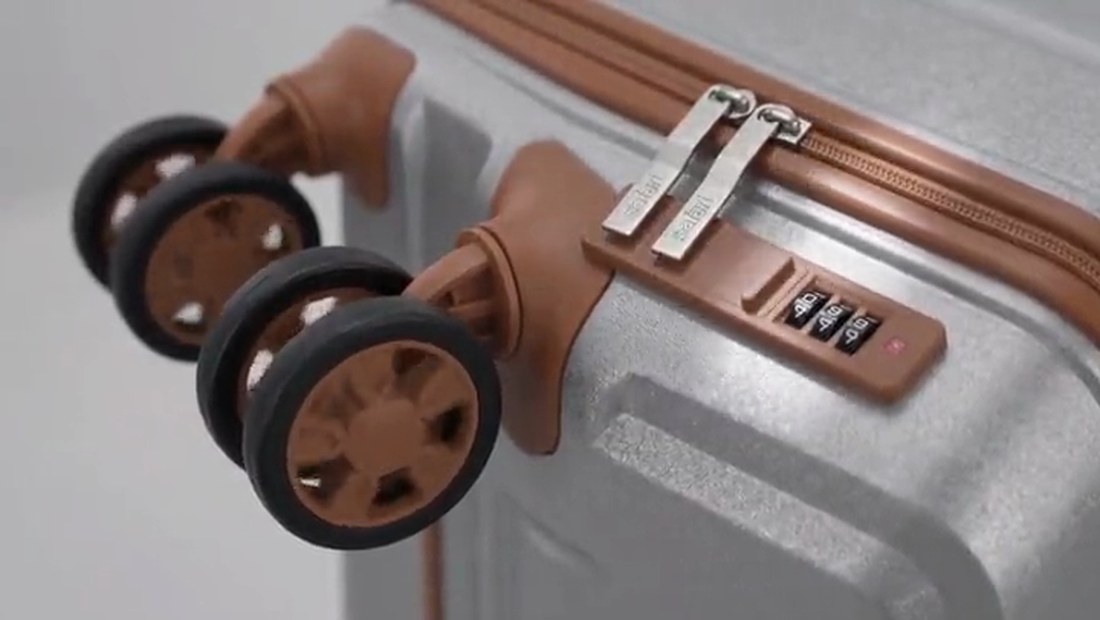}{frame 0:12} & {\scriptsize Borrow, generate or edit the scene} & {\scriptsize Suitcase, last scene (0:09-0:15). Two pairs of spinner wheels stick out of the case right beside the combination lock; on the real case the lock sits high on the side seam and the wheels are only at the four bottom corners. The scene judge flagged it as an AI issue but the product judge passed it; 1 of 1 reviewer marked the product wrong in an early pilot round.\enspace\href{https://github.com/anoubhav/ad-review-and-edit/blob/main/gallery/README.md\#g05}{\faImage}} \\
\midrule
\multicolumn{4}{@{}l}{\textbf{Scene quality}} \\* \addlinespace[1pt]
\textbf{Good scene (Keep)} & \gpic{3.00cm}{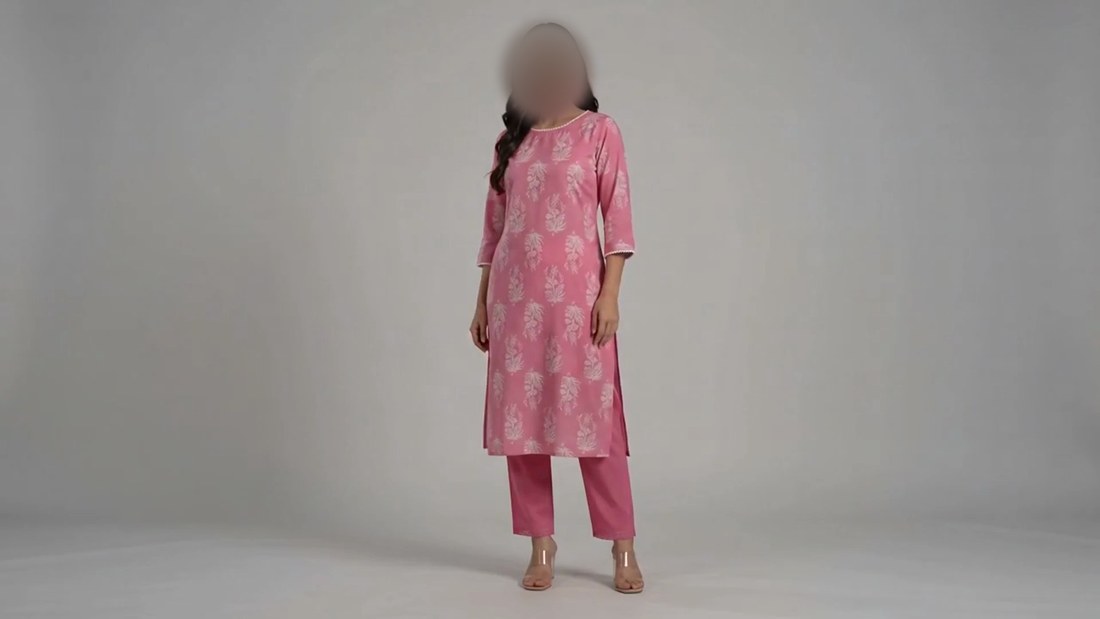}{frame 0:06.3 (start, wide)}\hspace{0.12cm}\gpic{3.00cm}{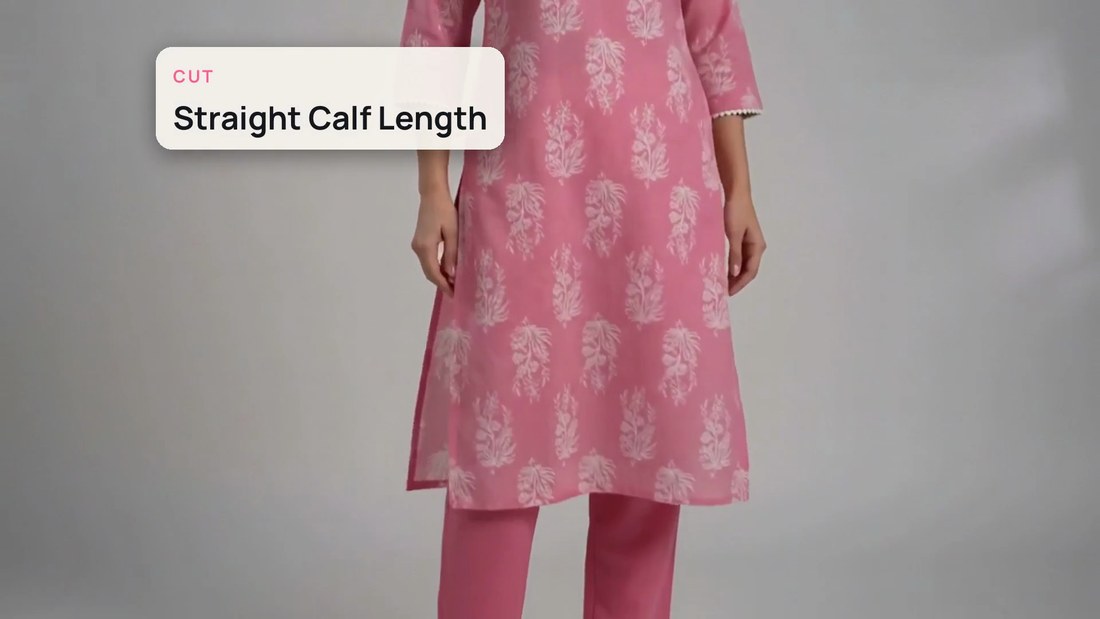}{frame 0:08.6 (end of the push-in)} & {\scriptsize Keep the scene} & {\scriptsize Kurta set, a good scene for contrast: the camera starts wide on the outfit and pushes in on the print and hem while a caption names the cut (0:06-0:09). The reviewer guide uses it as a Keep example (1 pilot reader marked Keep); the scene judge raised no finding on it.\enspace\href{https://github.com/anoubhav/ad-review-and-edit/blob/main/gallery/README.md\#g13}{\faImage}} \\
\addlinespace[3pt]
\textbf{Ghost copies overlap} & \gpic{3.00cm}{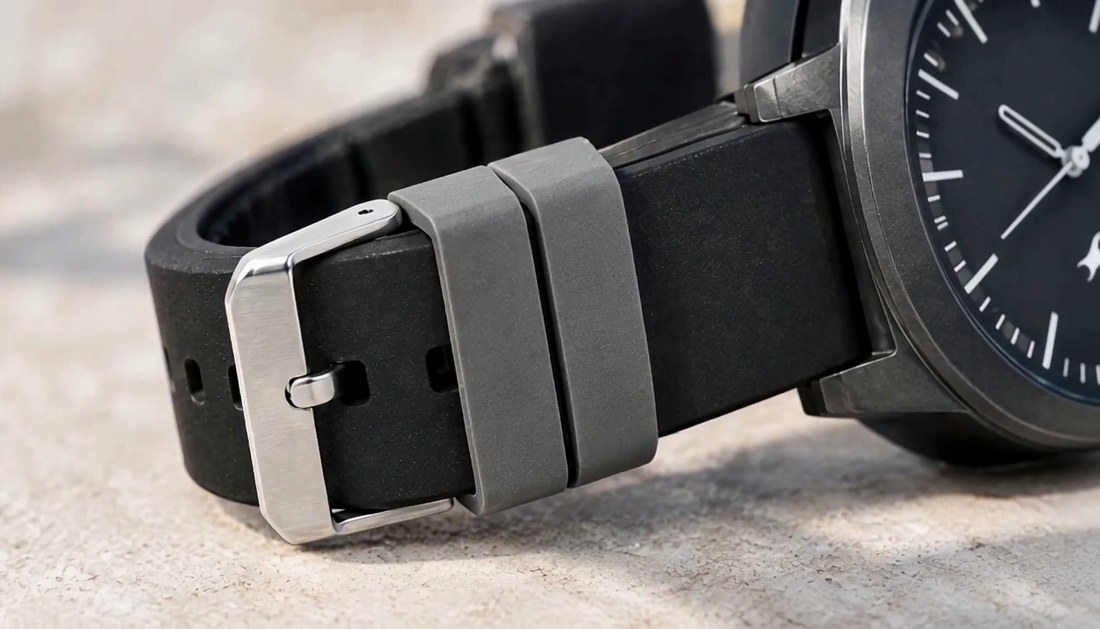}{frame 0:08.4 (just before)}\hspace{0.12cm}\gpic{3.00cm}{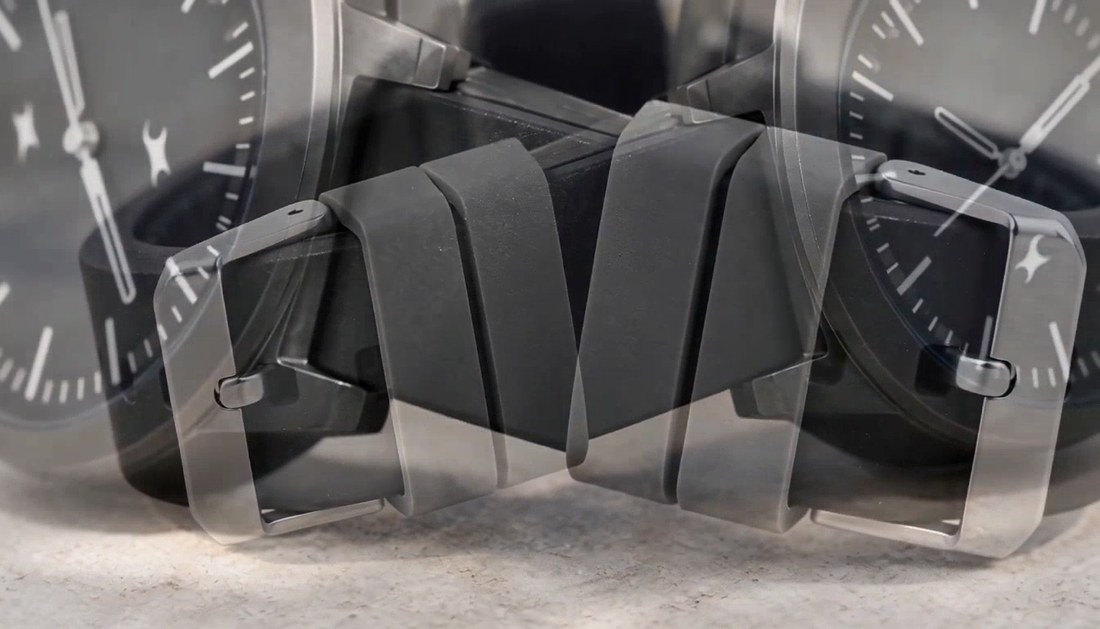}{frame 0:09.0 (copies overlap)} & {\scriptsize Borrow, generate or edit the scene} & {\scriptsize Wristwatch, strap close-up. Half a second into the scene, as the camera changes angle, several translucent copies of the watch face and buckle overlap, then settle into one strap (about 0:09). Judge flagged (frames judge accepted 3/3, clip judge 2/3); 3 of 3 reviewers marked swap (AI issue).\enspace\href{https://github.com/anoubhav/ad-review-and-edit/blob/main/gallery/README.md\#g14}{\faPlayCircle}} \\
\addlinespace[3pt]
\textbf{Crowded frame} & \gpic{3.00cm}{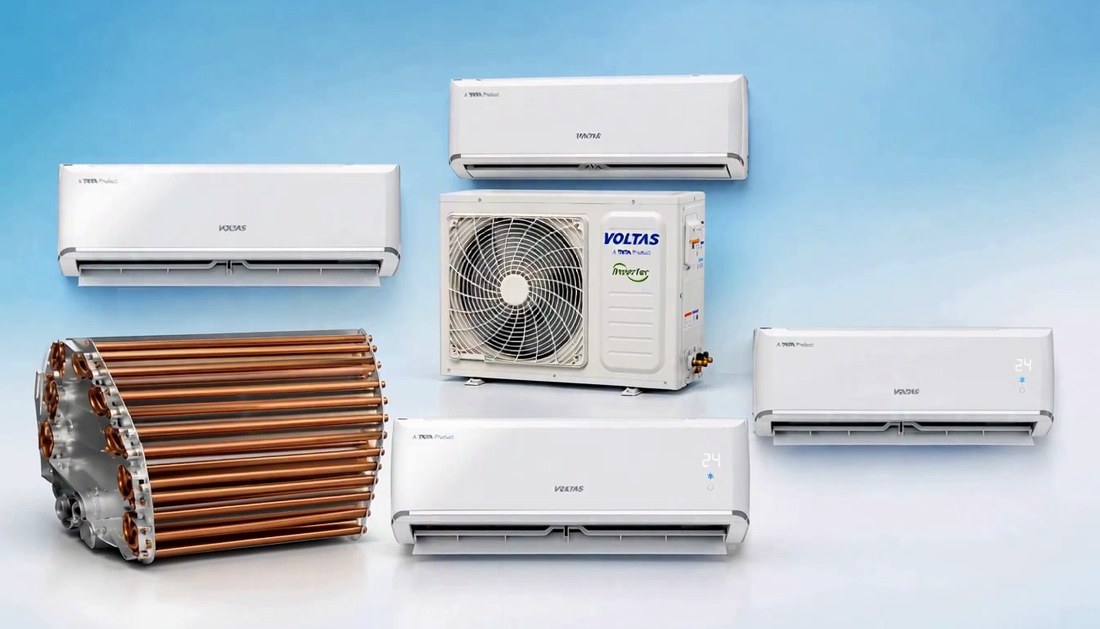}{frame 0:03.4}\hspace{0.12cm}\gpic{3.00cm}{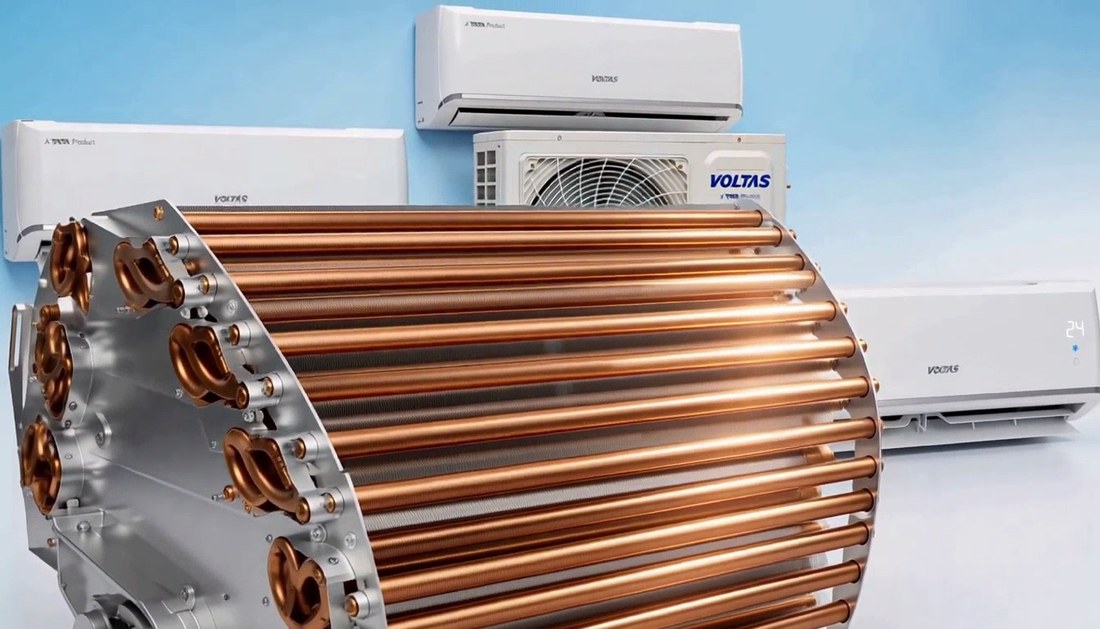}{frame 0:05.6} & {\scriptsize Borrow, generate or edit the scene} & {\scriptsize Air conditioner. A cluster of duplicated indoor units and an oversized copper coil fill the frame, and then the coil fills most of the frame, so the product is hard to read (0:03-0:07). Judge flagged (accepted 3/3); 3 of 3 reviewers marked swap (repetitive, dull or crowded).\enspace\href{https://github.com/anoubhav/ad-review-and-edit/blob/main/gallery/README.md\#g15}{\faPlayCircle}} \\
\addlinespace[3pt]
\textbf{Repeats the opening} & \gpic{3.00cm}{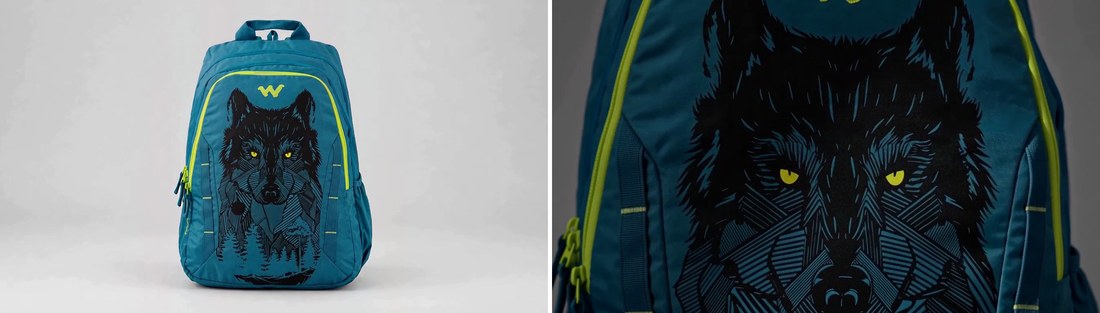}{opening scene: frames 0:00.2 and 0:03.7}\hspace{0.12cm}\gpic{3.00cm}{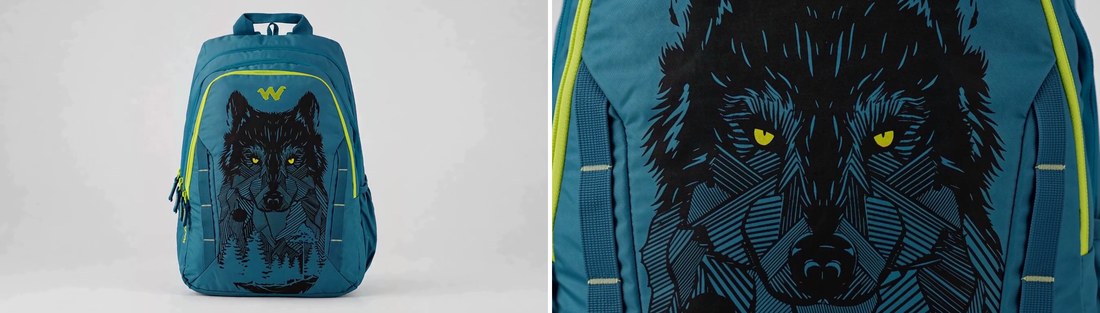}{closing scene: frames 0:11.7 and 0:15.0} & {\scriptsize Borrow, generate or edit the scene} & {\scriptsize Backpack. The closing scene repeats the opening scene: the same front-on framing and the same push-in to the printed panel (0:00-0:04 and 0:11-0:15). Both scene judges flagged it (each accepted 3/3); 3 of 3 reviewers marked the closing scene swap (repetitive, dull or crowded).\enspace\href{https://github.com/anoubhav/ad-review-and-edit/blob/main/gallery/README.md\#g16}{\faPlayCircle}} \\
\addlinespace[3pt]
\textbf{Set changes mid-ad} & \gpic{3.00cm}{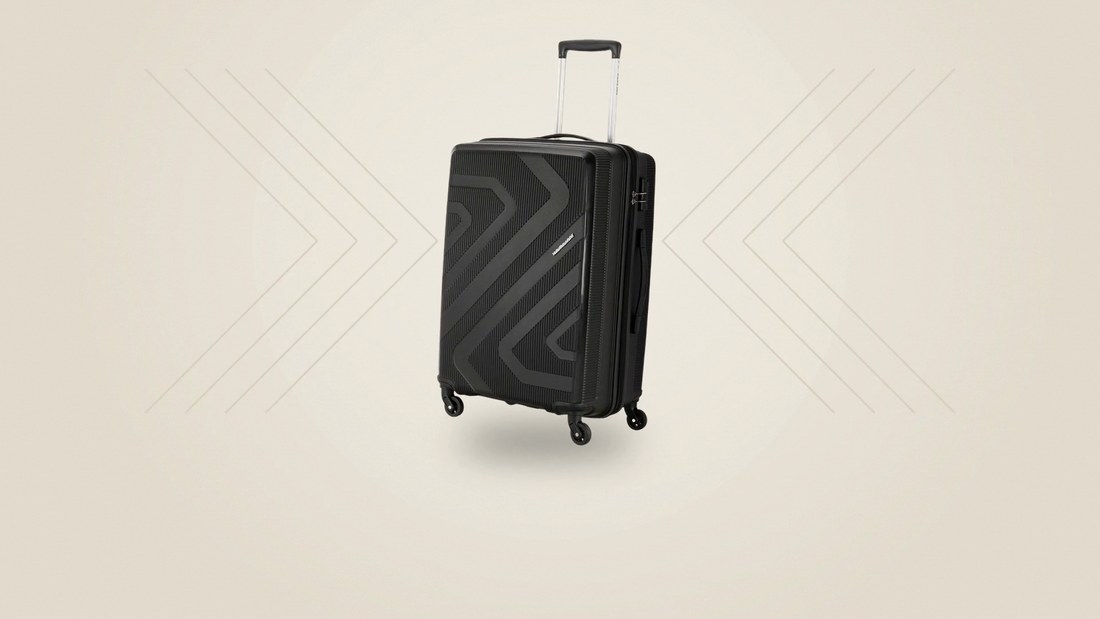}{frame 0:02.3 (scene 2)}\hspace{0.12cm}\gpic{3.00cm}{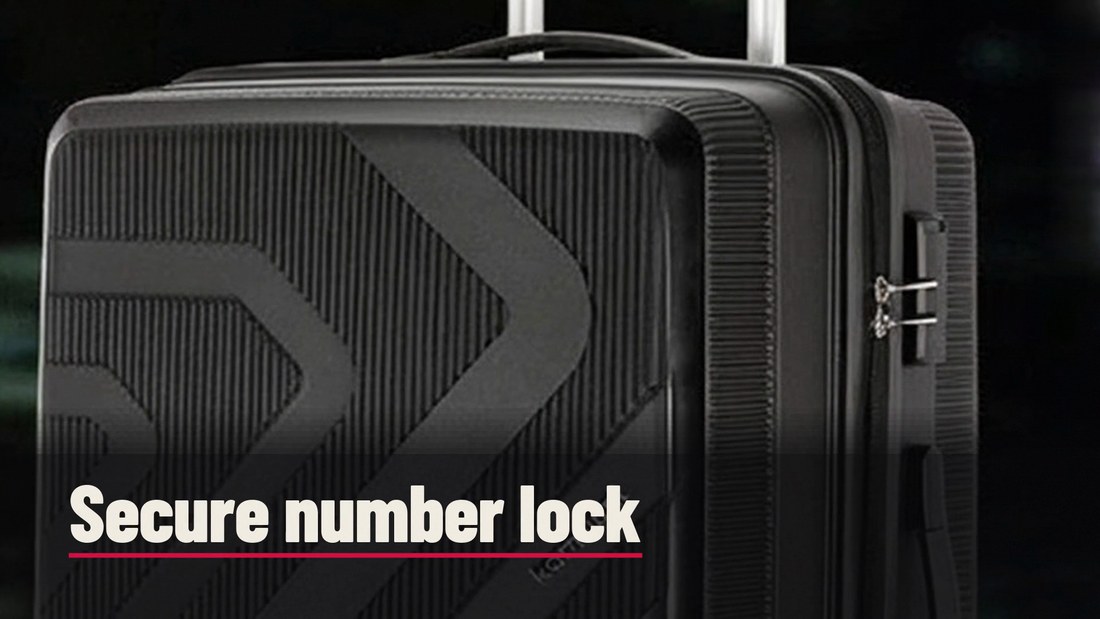}{frame 0:10.2 (scene 4)} & {\scriptsize Borrow, generate or edit the scene} & {\scriptsize Suitcase ad from a production run. Scenes 1 to 3 sit on a cream studio background; at 0:09.4 scene 4 cuts to a dark set, and the last scene goes back to cream. Judge flagged (accepted 1/1); 1 of 1 reviewer marked swap (world or continuity breaks).\enspace\href{https://github.com/anoubhav/ad-review-and-edit/blob/main/gallery/README.md\#g17}{\faImage}} \\
\addlinespace[3pt]
\textbf{Units multiply} & \gpic{1.96cm}{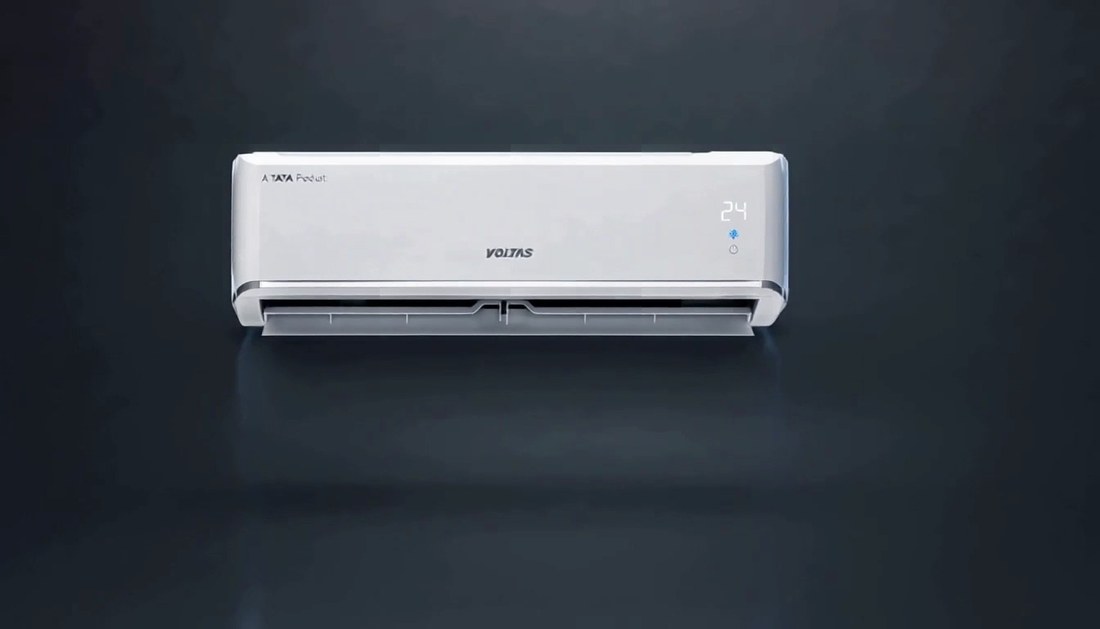}{frame 0:12.6}\hspace{0.12cm}\gpic{1.96cm}{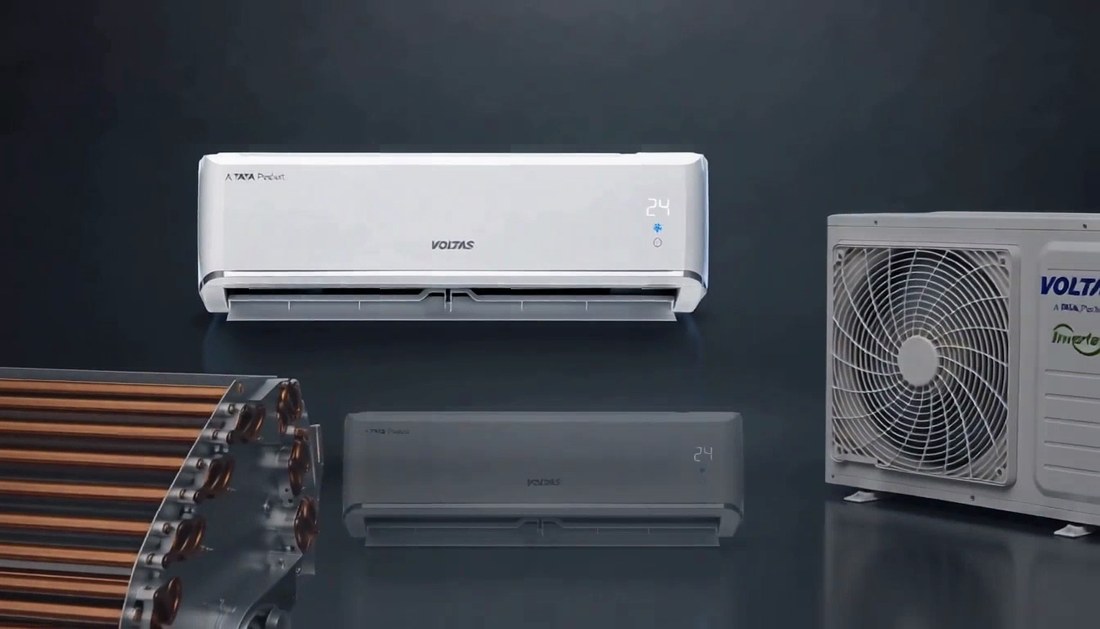}{frame 0:12.8}\hspace{0.12cm}\gpic{1.96cm}{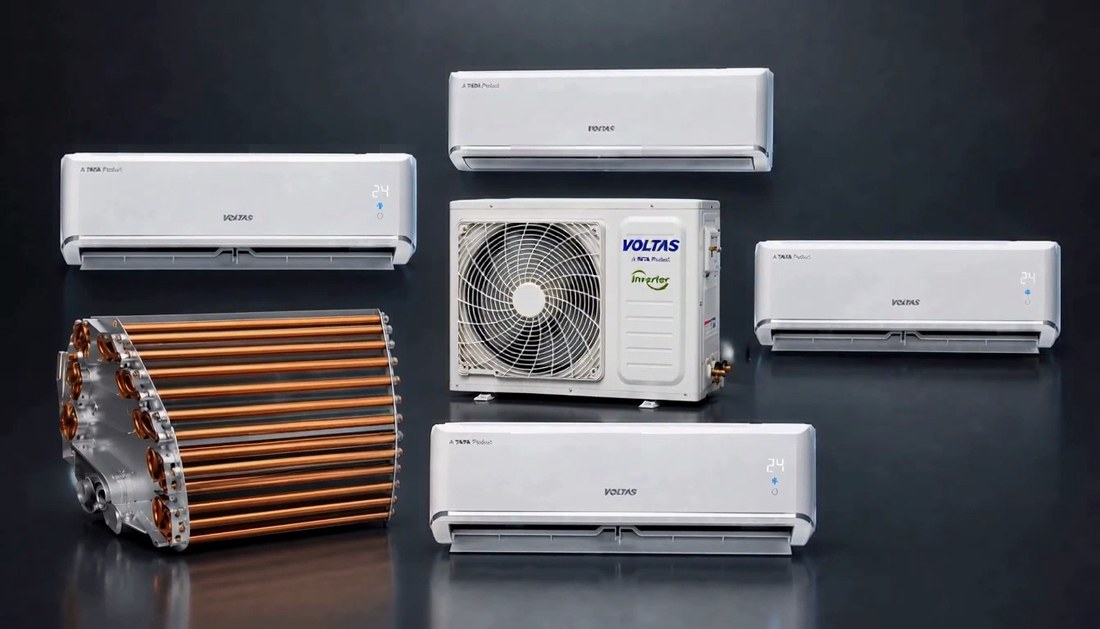}{frame 0:15.0 (end)} & {\scriptsize Borrow, generate or edit the scene} & {\scriptsize Air conditioner, closing scene. One indoor unit becomes four, plus an outdoor unit and a copper coil: they fade in from nowhere and drift into place (0:12.4-0:15). Clip judge flagged (accepted 3/3; frames judge 2/3); 3 of 3 reviewers marked swap (impossible motion).\enspace\href{https://github.com/anoubhav/ad-review-and-edit/blob/main/gallery/README.md\#g18}{\faPlayCircle}} \\
\addlinespace[3pt]
\textbf{Shoe floats through model} & \gpic{1.96cm}{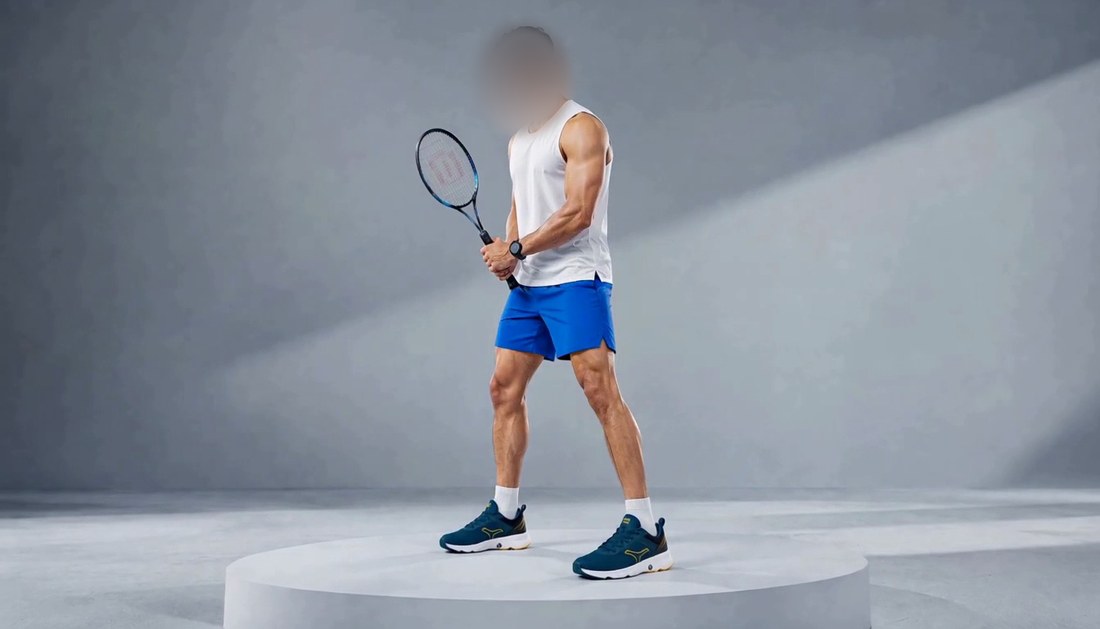}{frame 0:00.5}\hspace{0.12cm}\gpic{1.96cm}{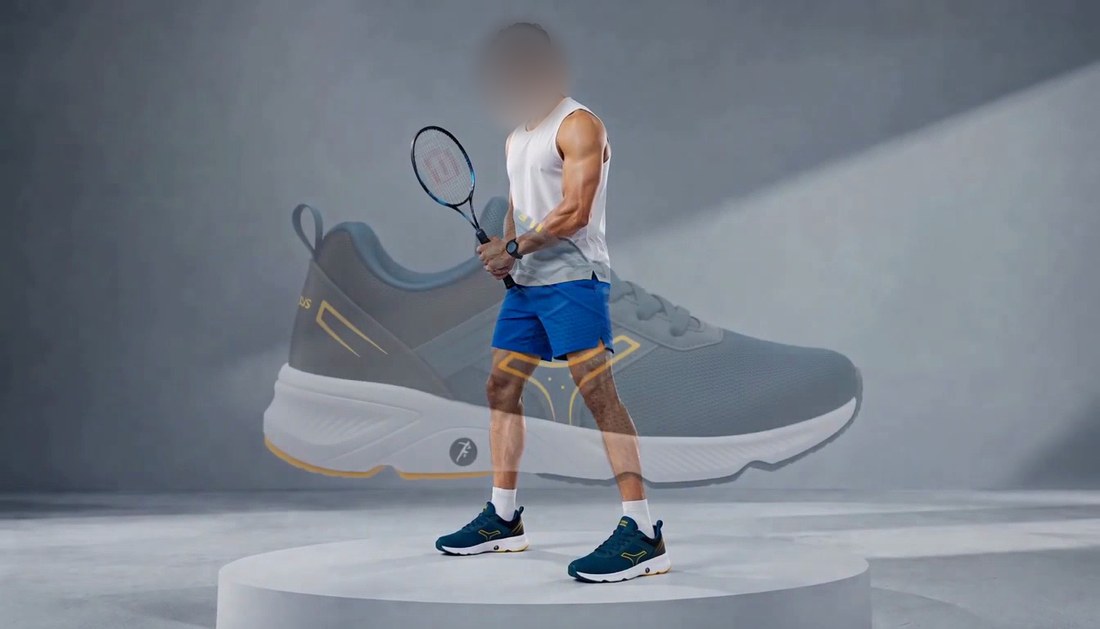}{frame 0:00.9}\hspace{0.12cm}\gpic{1.96cm}{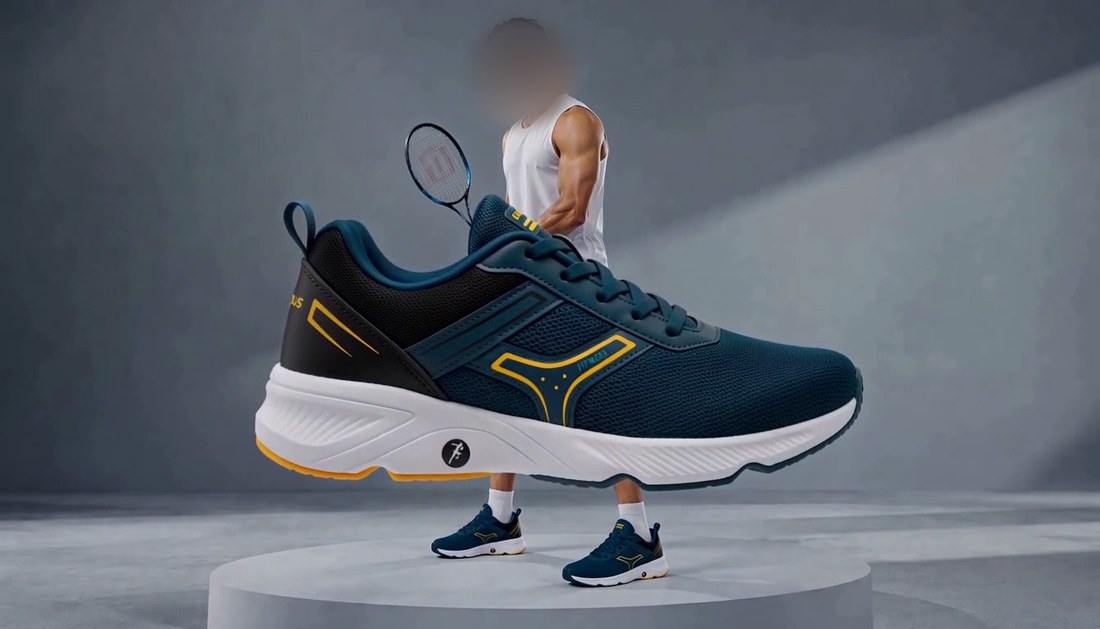}{frame 0:01.4} & {\scriptsize Borrow, generate or edit the scene} & {\scriptsize Sports shoe, opening scene. A giant translucent shoe fades in over the model and ends up floating in front of him, passing through his body (0:00.5-0:01.5). Both scene judges flagged it (each accepted 3/3); 3 of 3 reviewers marked swap (impossible motion).\enspace\href{https://github.com/anoubhav/ad-review-and-edit/blob/main/gallery/README.md\#g19}{\faPlayCircle}} \\
\addlinespace[3pt]
\textbf{Ends on a tight crop} & \gpic{3.00cm}{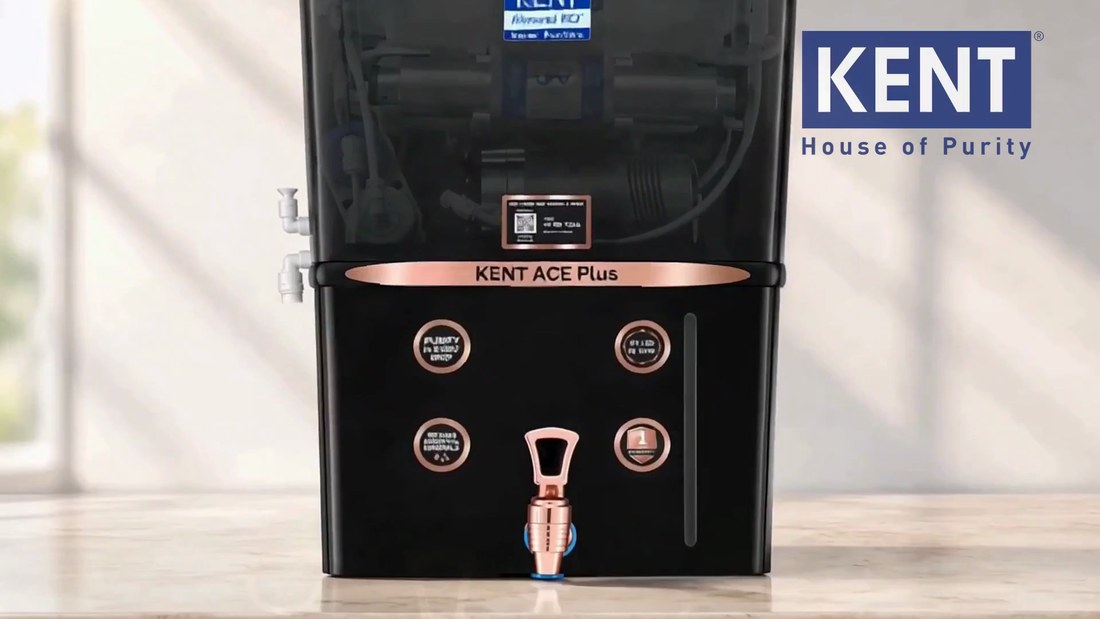}{frame 0:13.0}\hspace{0.12cm}\gpic{3.00cm}{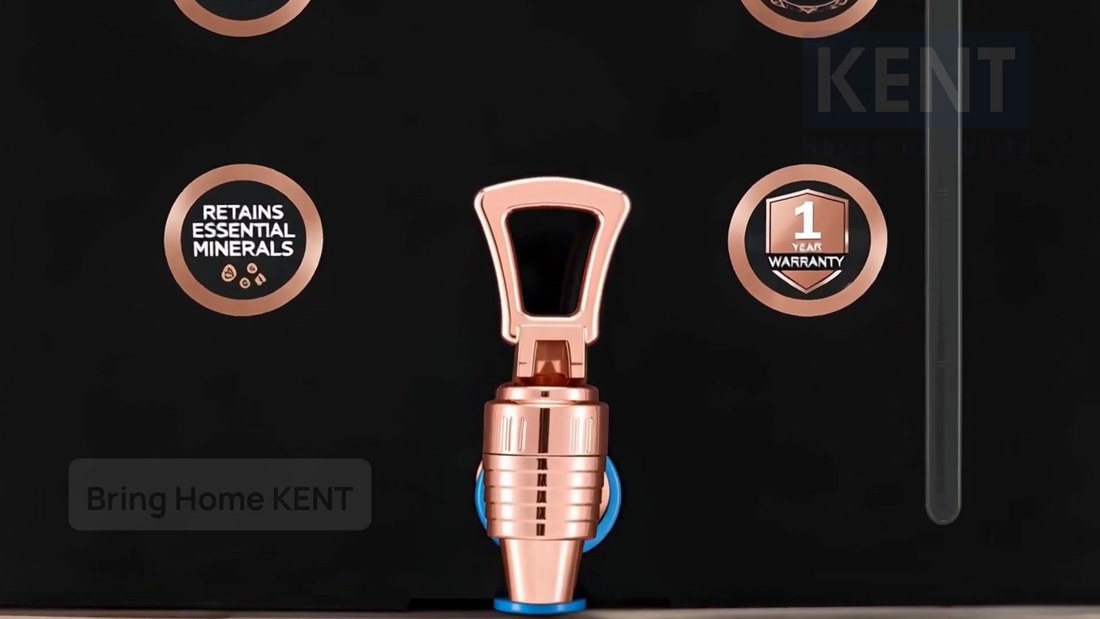}{frame 0:14.9 (last frame)} & {\scriptsize Borrow, generate or edit the scene} & {\scriptsize Water purifier, closing scene. The scene opens on the front of the purifier, then pushes in until the ad ends on a tight crop of the tap (last frame 0:14.9). Both scene judges flagged it (each accepted 3/3); 3 of 3 reviewers marked swap (weak ending).\enspace\href{https://github.com/anoubhav/ad-review-and-edit/blob/main/gallery/README.md\#g20}{\faPlayCircle}} \\
\midrule
\multicolumn{4}{@{}l}{\textbf{Pronunciation}} \\* \addlinespace[1pt]
\textbf{Word said wrong} & \gpic{3.43cm}{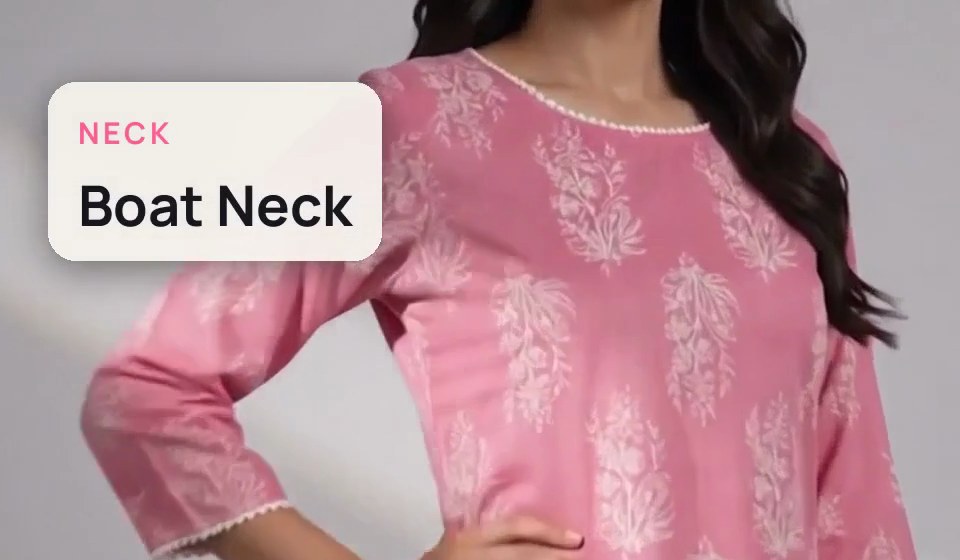}{frame, scene 2 (caption card)}\par\vspace{2pt}\gpic{6.11cm}{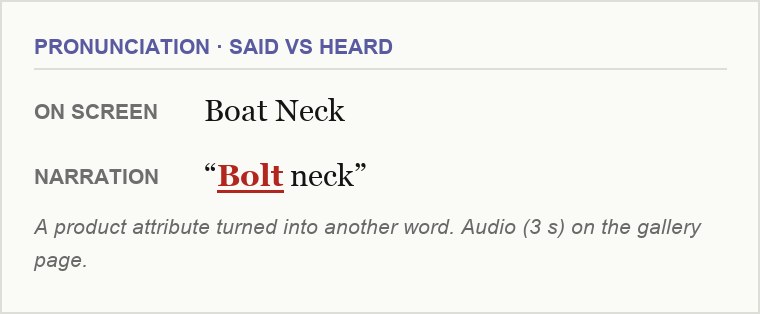}{} & {\scriptsize Respell the word and re-voice the line} & {\scriptsize Kurta set, scene 2 of 4: the on-screen card reads `Boat Neck', but the narrator says `bolt neck'. 1 reviewer marked it in an early pilot round; no judge result is recorded for it.\enspace\href{https://github.com/anoubhav/ad-review-and-edit/blob/main/gallery/README.md\#g22}{\faVolumeUp}} \\
\addlinespace[3pt]
\textbf{Respelled, re-voiced} & \gpic{6.11cm}{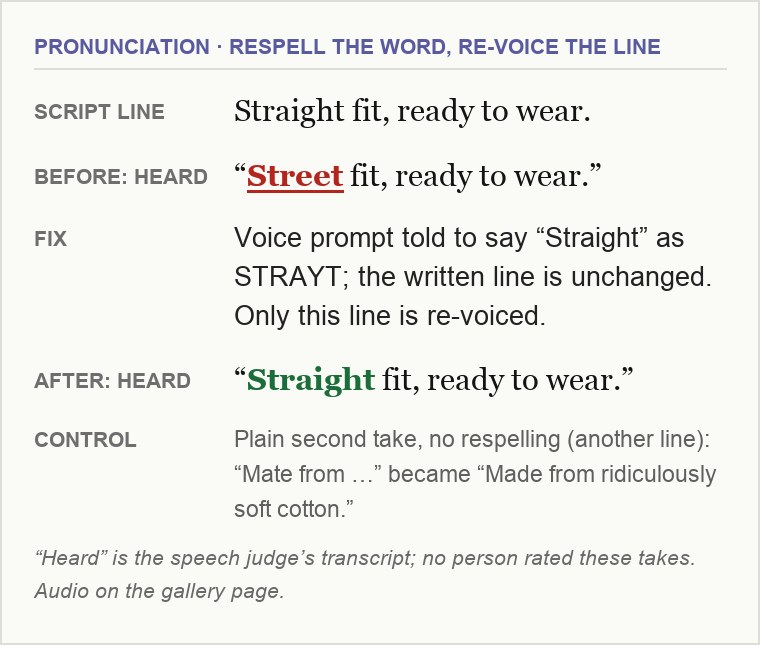}{} & {\scriptsize\textit{Word respelled, line re-voiced}} & {\scriptsize Trousers ad, line `Straight fit, ready to wear.': the speech judge heard `Street fit'. After the word was respelled STRAYT in the voice prompt and only that line was re-voiced, it heard `Straight fit'. Only the speech judge checked the two takes; no person did. Control: a plain second take with no respelling turned `Mate from' into `Made from' on another line, and across the whole test respelling fixed about as many flagged lines as a plain second take (27\% vs 25\%).\enspace\href{https://github.com/anoubhav/ad-review-and-edit/blob/main/gallery/README.md\#g23}{\faVolumeUp}} \\
\midrule
\multicolumn{4}{@{}l}{\textbf{Script}} \\* \addlinespace[1pt]
\textbf{Claim not in the source} & \gpic{3.55cm}{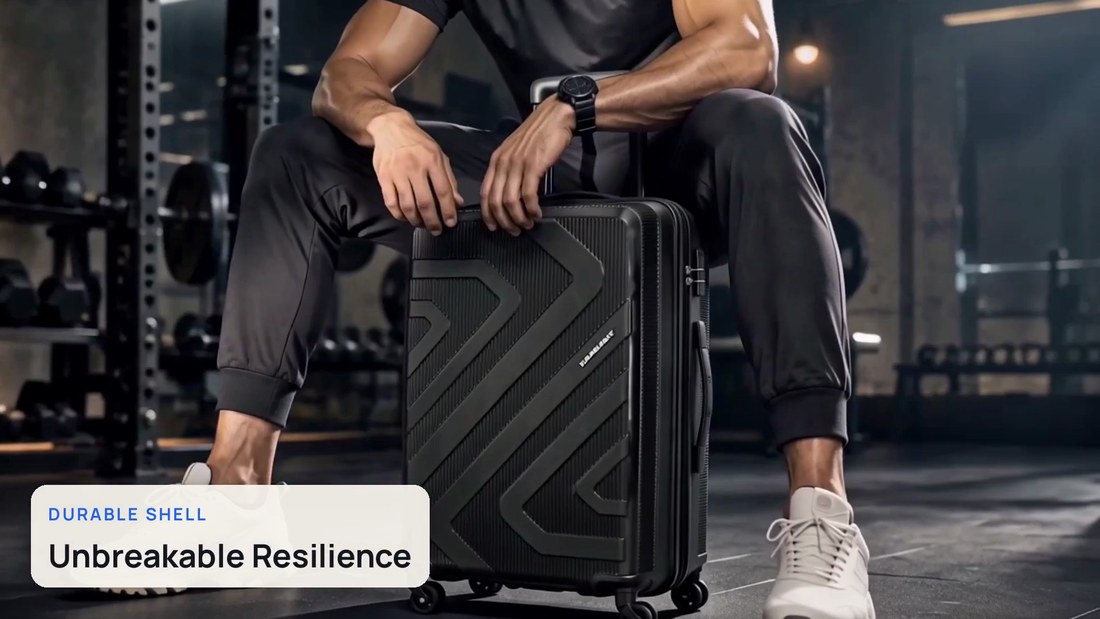}{frame 0:01.5}\par\vspace{2pt}\gpic{6.11cm}{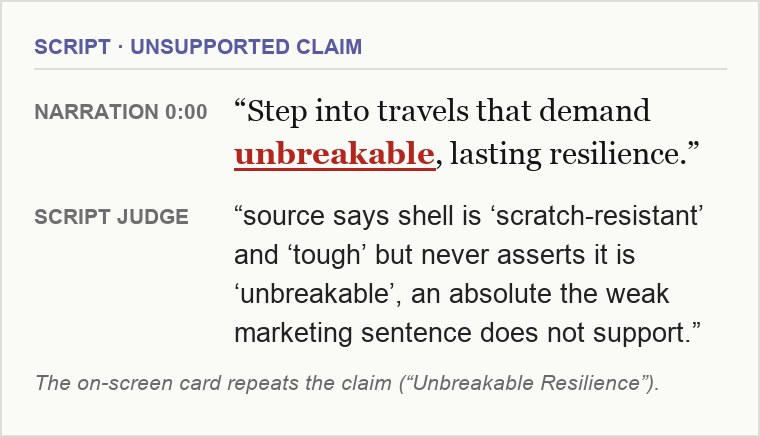}{} & {\scriptsize Correct the claim or regenerate the line} & {\scriptsize Suitcase ad, 0:00-0:04: the narration says the shell is `unbreakable', but the product record only calls it scratch-resistant and tough. The script judge flagged it, and all 3 reviewers marked it by accepting that finding, so the marks are not independent of the judge.\enspace\href{https://github.com/anoubhav/ad-review-and-edit/blob/main/gallery/README.md\#g24}{\faVolumeUp}} \\
\addlinespace[3pt]
\textbf{Number without its unit} & \gpic{3.55cm}{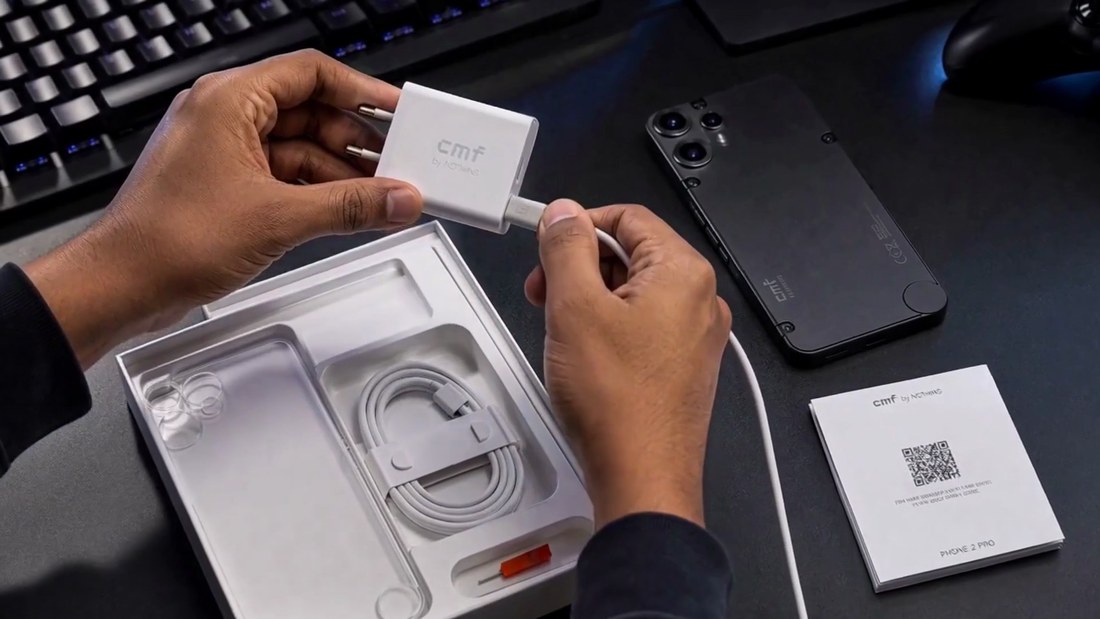}{frame 0:07}\par\vspace{2pt}\gpic{6.11cm}{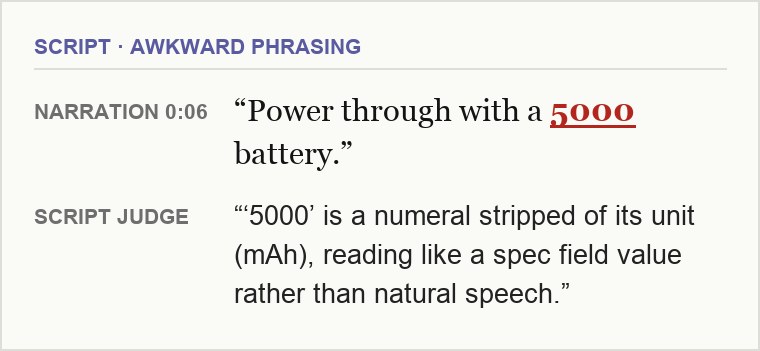}{} & {\scriptsize Rephrase or regenerate the line} & {\scriptsize Smartphone ad from a production review outside the round, 0:06: the line `Power through with a 5000 battery.' drops the unit (mAh), so it sounds like a spec-sheet field. The script judge flagged it, and so did a second script judge that this review also ran and that was later dropped (each finding got 1 thumbs-up and none down). Both reviewers marked awkward phrasing: one on this line, one for the ad as a whole.\enspace\href{https://github.com/anoubhav/ad-review-and-edit/blob/main/gallery/README.md\#g25}{\faVolumeUp}} \\
\addlinespace[3pt]
\textbf{Claim narrowed} & \gpic{6.11cm}{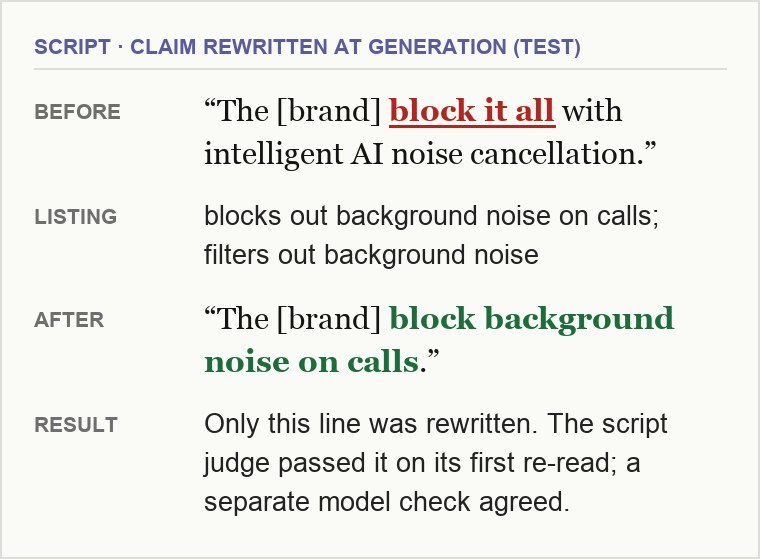}{} & {\scriptsize\textit{Line regenerated at generation}} & {\scriptsize Earbuds ad from the 410-ad script-rewrite test of Sec. 5: `The [brand] block it all with intelligent AI noise cancellation.' claims more than the listing, which only says they block background noise on calls and filter out background noise. Only that line was rewritten at generation time, as `The [brand] block background noise on calls.' The script judge flagged the old line and passed the new one on its first re-read, and a separate model check agreed. The script is one of the 129 the annotators then read cold, and no annotator flagged it.\enspace\href{https://github.com/anoubhav/ad-review-and-edit/blob/main/gallery/README.md\#g27}{\faImage}} \\
\midrule
\multicolumn{4}{@{}l}{\textbf{Sound}} \\* \addlinespace[1pt]
\textbf{Wrong accent} & \gpic{6.11cm}{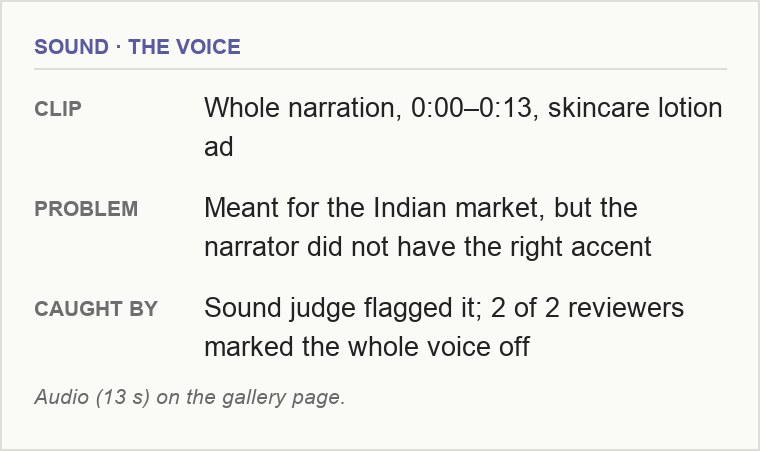}{} & {\scriptsize Re-voice the ad} & {\scriptsize Skincare lotion ad from a production run before the round, whole narration (0:00-0:13): the ad was meant for the Indian market, but the narrator did not have the right accent. The sound judge flagged it (in the round it flagged no ad, Sec. 6), and both reviewers marked the whole voice as off after accepting its finding.\enspace\href{https://github.com/anoubhav/ad-review-and-edit/blob/main/gallery/README.md\#g28}{\faVolumeUp}} \\
\addlinespace[3pt]
\textbf{Voice swapped} & \gpic{6.11cm}{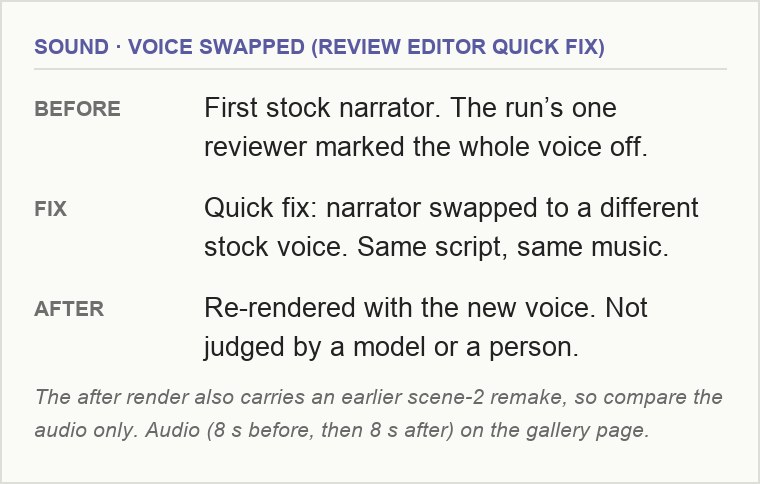}{} & {\scriptsize\textit{Voice swapped}} & {\scriptsize Eyeliner ad from a production run, whole narration: the run's one reviewer marked the whole voice off, and a Quick fix in the review editor swapped in a different stock voice, keeping the same script and music. This is a plain voice swap, separate from the speech-to-speech re-voicing of Tab. 3. No judge flagged the voice, and no model or person judged the new render. The new render also has an earlier scene-2 remake, so compare the audio only.\enspace\href{https://github.com/anoubhav/ad-review-and-edit/blob/main/gallery/README.md\#g29}{\faVolumeUp}} \\
\midrule
\multicolumn{4}{@{}l}{\textbf{Captions}} \\* \addlinespace[1pt]
\textbf{Card over the wheel} & \gpic{3.55cm}{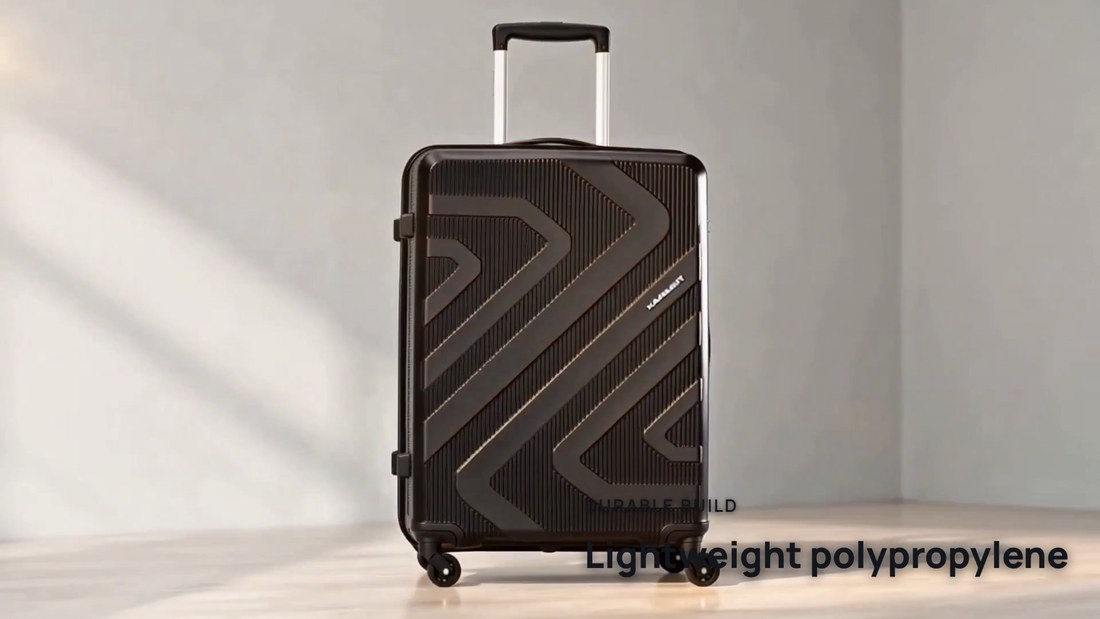}{frame 0:02} & {\scriptsize Move or restyle the caption} & {\scriptsize Suitcase ad, 0:00-0:04: the caption sits on the wheel and bottom corner of the suitcase even though the frame has empty space around it. 3 of 3 reviewers marked placement; a placement judge, run after the round and never shown to reviewers, flagged it in both of its runs.\enspace\href{https://github.com/anoubhav/ad-review-and-edit/blob/main/gallery/README.md\#g30}{\faPlayCircle}} \\
\addlinespace[3pt]
\textbf{Card over a face} & \gpic{3.55cm}{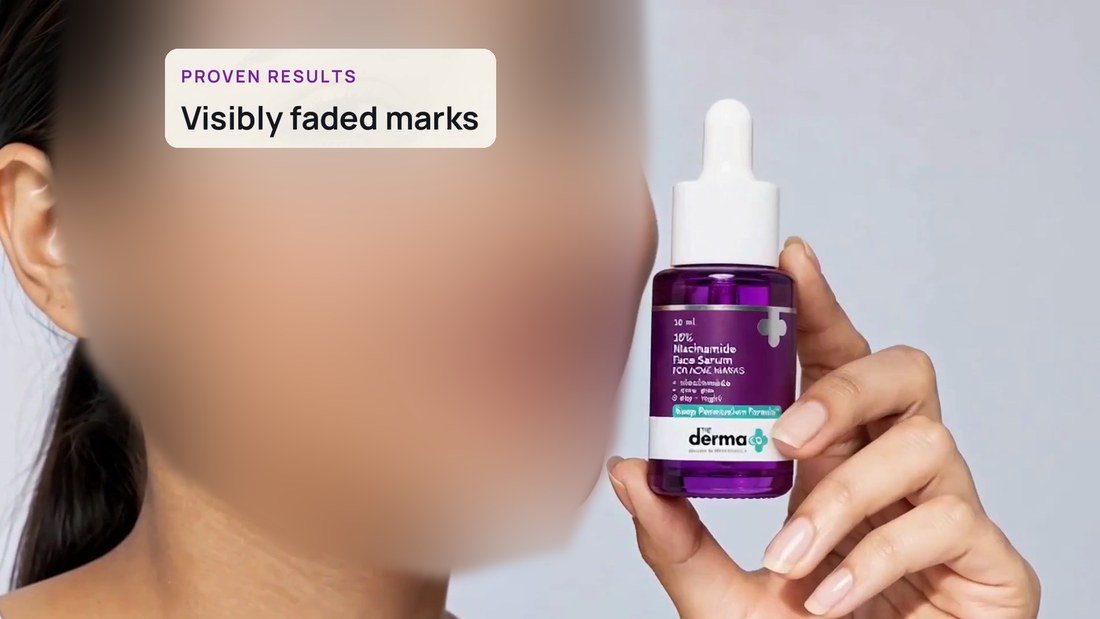}{frame 0:11 (face blurred)} & {\scriptsize Move or restyle the caption} & {\scriptsize Face serum ad, 0:10-0:12: the callout card sits on the model's forehead (face blurred here, card left as rendered). 2 of 3 reviewers marked placement; the placement judge, run after the round and never shown to reviewers, flagged it in both of its runs.\enspace\href{https://github.com/anoubhav/ad-review-and-edit/blob/main/gallery/README.md\#g31}{\faPlayCircle}} \\
\addlinespace[3pt]
\textbf{Caption about a hidden part} & \gpic{3.55cm}{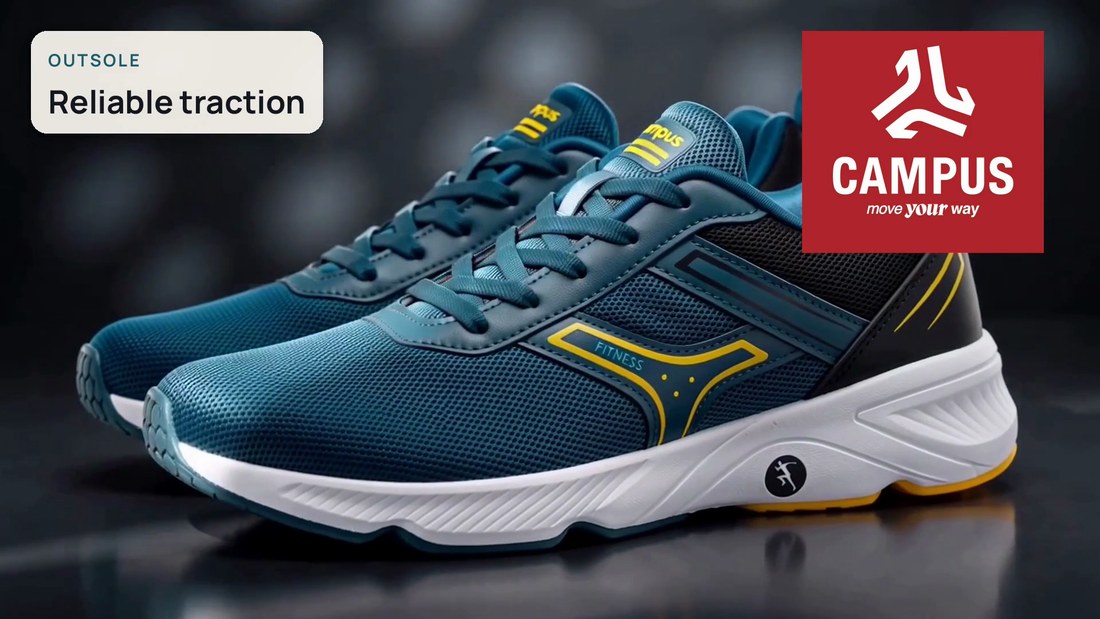}{frame 0:13} & {\scriptsize Reword or remove the caption} & {\scriptsize Sports shoe ad, 0:12-0:15: the card reads `Outsole: Reliable traction', but the shoes stand flat and the sole's tread is never shown. 2 of 3 reviewers marked relevance; the relevance judge, run after the round and never shown to reviewers, flagged it in both of its runs.\enspace\href{https://github.com/anoubhav/ad-review-and-edit/blob/main/gallery/README.md\#g32}{\faPlayCircle}} \\
\addlinespace[3pt]
\textbf{Overlay restyled (unshipped prototype)} & \gpic{3.91cm}{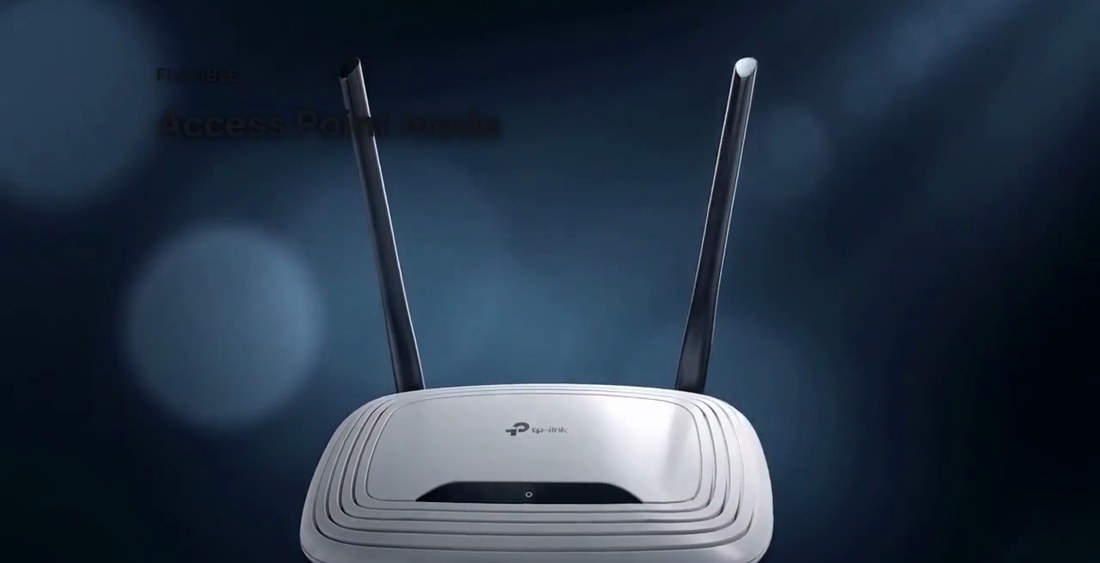}{before: as delivered, 0:12} & \gpic{2.91cm}{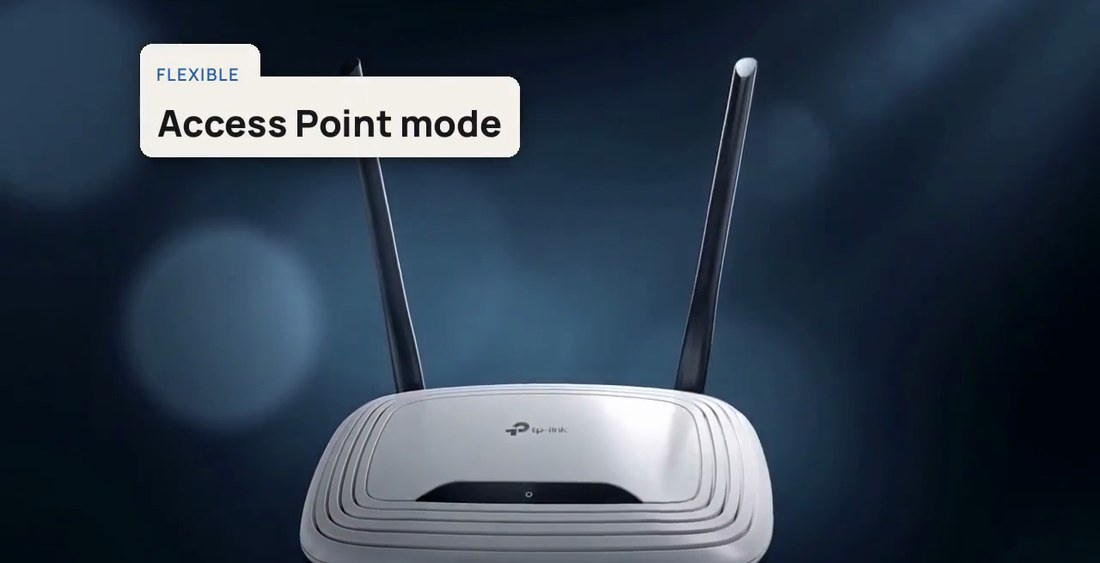}{after: prototype restyle (never shipped)}\par\vspace{1pt}{\scriptsize\textit{Caption restyled (prototype)}} & {\scriptsize Router ad, 0:12: the delivered caption is near-black text on a navy background and cannot be read. Restyling only the caption layer put the same words on a cream card without touching the video. This is an unshipped prototype, and only the project owner looked at the after. Reviewers are not asked about readability, but 2 of 3 marked this caption's placement; the placement judge passed it in both runs.\enspace\href{https://github.com/anoubhav/ad-review-and-edit/blob/main/gallery/README.md\#g33}{\faImage}} \\
\midrule
\end{longtable}}

\section{The Same Ads with Judges Hidden}
\label{app:alone}

R1, a reviewer who had not seen the ads, reviewed the round's ads in a separate run of the portal with every judge hidden. The run held 60 of the 120 ads, three per product, in an order that cycles through the 20 products; every question but sound was required and the secondary options were offered (a blank sound answer counts as no issue; sound was left blank on every ad). The units, the rules that turn answers into marks (those of the round), the outcomes, the bootstrap and the exclusions were written down and the analysis code hashed before the first review; one amendment, made before any review was submitted, records the scene-quality confound below. R1 submitted all 60 ads. Checks: every judge was hidden, the run holds no thumbs, and the stored findings on the 60 ads are those of the round (360 of 360 ad--judge pairs unchanged). Pronunciation was computed but is not reported, because the narration engine and its judge changed after the round.

\Cref{tab:alone} compares, on the same ads and units, each reviewer alone, the stored judge findings and the round's reviewers, who saw them. A unit is marked as in the round: a scene with the product marked wrong or a swap the reviewer chose, a line marked bad, a caption marked for placement or relevance. Intervals are percentiles of 2{,}000 bootstrap resamples of ads. $^\dagger$In the round, marking a scene's product wrong pre-filled a swap for it, and an older grid hid such scenes from the scene-quality question, so the round was asked about fewer scenes (in parentheses) and its scene-quality marks mix the reviewer, the findings and the pre-fill.

Time is descriptive, since the two groups have different reviewers. Per-question gaps over 15 minutes count as breaks and are dropped; per-review times run from the first answer to submission, dropping reviews over 4 hours (one of R1's). R1 took a median of 3.4 [2.4, 4.7] minutes from first answer to submission (quartiles in brackets; 59 reviews), against 4.1 [2.8, 6.6] for the round's 139 reviews of the same ads. Median time per question, R1 against the round: scene quality 44\,s against 0\,s; script 43\,s against 65\,s; caption placement 30\,s against 7\,s; caption relevance 1\,s against 2\,s. The product is answered first, and the review record starts at the first answer, so its time cannot be measured; in the round the product answer pre-filled the scene-quality grid, so its scene-quality time is near zero.

R2, one of the round's reviewers, later reviewed all 60 ads in the same run with every judge hidden, under a second amendment written before any of these reviews was read. R2 had also reviewed the same 60 ads in the round about two weeks earlier, with judge findings on screen, and the amendment pre-specified the comparison of the two passes. On the same scenes, the product was marked wrong on 57\% [49, 65] with judge findings and 19\% [13, 26] with judges hidden ($\kappa$ 0.22 between the passes); of the 7 lines the script judge flagged, 7 were marked with its findings on screen and 1 without. The pass with judges hidden came second, so memory of the findings would, if anything, narrow these gaps. A review by R2 took a median of 2.3 [1.5, 4.4] minutes with judges hidden (59 timed reviews; quartiles in brackets), against 3.0 [2.3, 4.2] with judge findings in the round; this was R2's second look at these ads, so the shorter time cannot be credited to hiding the judges. Pooled over both reviewers, a review with judges hidden took a median of 2.9 minutes (118 timed reviews). The two reviewers working alone agreed, on the 60 ads both reviewed, at Cohen's $\kappa$ product 0.17 [0.02, 0.31]; scene quality 0.14 [0.04, 0.24]; script $-$0.04 [$-$0.05, $-$0.02]; caption placement 0.06 [$-$0.11, 0.25]; caption relevance $-$0.02 [$-$0.03, 0.00].

\paragraph{Over the session.} \Cref{sec:portal} gives R2's product marks over the first, middle and last 20 ads by submission time, with judges hidden and with judge findings, and R2's review times with judges hidden (\cref{fig:ww}b); on the same ads the round's majority marked 56\%, 54\% and 62\%. Against the round's majority on each ad, the fall in R2's marks with judges hidden has Spearman $\rho$ $-$0.27 over submission order (permutation $p$ 0.04); total marks per ad have $\rho$ $-$0.56 and minutes per review $-$0.59 (both $p$ $<$ 0.001). R1's product marks held (31\%, 37\% and 30\%; $\rho$ $-$0.06, $p$ 0.65) while reviews sped up from 4.3 to 2.5 minutes ($\rho$ $-$0.45, $p$ $<$ 0.001). This analysis was added after the results were read.

\begin{table}[h]
\centering\small
\caption{The same ads with judges hidden (60 ads each). Alone: share of units the reviewer marked with every judge hidden. Judge: share the stored findings flagged. Alone's marks flagged: of the units the reviewer marked, the share the judge had flagged. Flags marked alone: of the judge's flags, the share the reviewer marked. Both columns give counts when fewer than 20. Round any and majority: share marked by any round reviewer, or by more than half of those asked, with judge findings on screen. $\kappa$: Cohen's $\kappa$ between the reviewer alone and the round's majority on units both were asked. Shares in \%, brackets are 95\% intervals. R2 was also one of the round's reviewers on these ads (the only one on 20 of them), so the round columns include that reviewer's own review and their $\kappa$ against the majority is not given.}
\label{tab:alone}
\setlength{\tabcolsep}{2.6pt}
\begin{tabular}{@{}llrrrrrrrr@{}}
\toprule
\hd{Axis} & \hd{Unit} & \hd{Units} & \hd{Alone} & \hd{Judge} & \hd{\begin{tabular}[b]{@{}r@{}}Alone's marks\\flagged\end{tabular}} & \hd{\begin{tabular}[b]{@{}r@{}}Flags marked\\alone\end{tabular}} & \hd{\begin{tabular}[b]{@{}r@{}}Round\\any\end{tabular}} & \hd{\begin{tabular}[b]{@{}r@{}}Round\\maj.\end{tabular}} & \hd{\begin{tabular}[b]{@{}r@{}}$\kappa$ alone\\vs maj.\end{tabular}} \\
\midrule
\multicolumn{10}{@{}l}{\emph{R1 (60 ads)}} \\
Product & scene & 280 & 33 [25, 41] & 68 & 74 [61, 85] & 36 [27, 44] & 67 & 58 & 0.11 [$-$0.02, 0.22] \\
\zebra Scene quality$^\dagger$ & scene & 280 (229) & 47 [38, 55] & 25 & 27 [20, 33] & 49 [37, 61] & 37 & 33 & 0.05 [$-$0.07, 0.17] \\
Script & line & 280 & 3 [1, 4] & 3 & 0 of 7 & 0 of 7 & 15 & 5 & $-$0.03 [$-$0.05, $-$0.02] \\
\zebra Caption placement & caption & 255 & 13 [8, 19] & no judge &  &  & 17 & 5 & 0.06 [$-$0.06, 0.21] \\
Caption relevance & caption & 255 & 5 [2, 8] & no judge &  &  & 6 & 2 & $-$0.03 [$-$0.05, $-$0.01] \\
\midrule
\multicolumn{10}{@{}l}{\emph{R2 (60 ads)}} \\
Product & scene & 280 & 19 [13, 26] & 68 & 93 [85, 98] & 26 [18, 34] & 67 & 58 &  \\
\zebra Scene quality$^\dagger$ & scene & 280 (229) & 31 [26, 36] & 25 & 41 [31, 51] & 49 [37, 62] & 37 & 33 &  \\
Script & line & 280 & 5 [3, 8] & 3 & 1 of 15 & 1 of 7 & 15 & 5 &  \\
\zebra Caption placement & caption & 255 & 16 [10, 21] & no judge &  &  & 17 & 5 &  \\
Caption relevance & caption & 255 & 1 [0, 3] & no judge &  &  & 6 & 2 &  \\
\bottomrule
\end{tabular}
\end{table}

\section{Baselines in Detail}
\label{app:baselines}

\Cref{tab:baselines} lists the primary score of every product-fidelity baseline of \cref{tab:experiments} on the 120 blind-labeled scenes (29 defective, 35 ads), with 95\% intervals from the same ad-clustered bootstrap (2{,}000 resamples) and the paired gap to the deployed clip judge. Lower predicted quality or higher dissimilarity counts as more likely defective; every sign and primary score was fixed before scoring and no threshold was tuned. Dropping the split-vote scenes (\cref{tab:labelsets}) likely removes hard cases, and the clip judge's prompt counts any confirmed difference as wrong, stricter than annotators may have been, so label noise can move these scores either way. Raising the clip judge's own threshold trades recall for precision: the first threshold at which at least 70\% of its flags are defective keeps 21 flags holding 15 of the 29 defective scenes, and the first at 85\% keeps 8 holding 7. DOVER~\cite{wu2023dover} and VideoScore2~\cite{he2025videoscore2} never see the catalog photos, so they cannot tell a wrong product from the right one; they show that a standard quality score would pass such scenes (VideoScore2 gives 107 of the 120 clips its top visual-quality score). The open judges received the deployed prompt, photos and output schema; Qwen3.5-9B~\cite{qwen2026qwen35} ran in bf16 and Qwen3.6-27B~\cite{qwen2026qwen36} in FP8, with thinking on as the primary setting (sampled at the temperature their model cards recommend, one sample per scene; thinking off was decoded greedily). A repeat with another seed and server layout moved the 9B's AUC by up to 0.09, so single rows carry that much run-to-run noise beyond the interval shown. The best open setting, the 27B with thinking off on eight frames, stays 0.12 below the clip judge, with an interval that includes zero, the only such gap besides the deployed judge's own eight-frame run; its verdicts reach $\kappa$ 0.35 (recall 17 of 29, precision 17 of 36), against 0.43 (26 of 29, 26 of 56) for the clip judge. With thinking on, the open models flag 69 to 99 of the 120 scenes ($\kappa$ 0.05 to 0.15). The prompt was written for the closed judge, and no variant was tried for the open ones. The clip judge's earlier runs and the starting-frame judge did not record their model version, so the eight-frame and starting-frame gaps may include a model change; the eight-frame run keeps the deployed prompt, while the starting-frame judge has its own (\cref{tab:experiments}).

\begin{table}[h]
\centering\small
\caption{Product-fidelity baselines on the 120 blind-labeled scenes. Gap: clip judge minus baseline, paired, with 95\% interval; intervals are not corrected for multiple comparisons.}
\label{tab:baselines}
\begin{tabular}{@{}l l r l l@{}}
\toprule
\hd{Score} & \hd{Input} & \hd{AUC} & \hd{95\% interval} & \hd{Gap to clip judge} \\
\midrule
Clip judge (deployed) & clip, catalog photos & 0.84 & [0.73, 0.94] &   \\
Starting-frame judge & starting frame, photos & 0.69 & [0.61, 0.78] & +0.15 [0.01, 0.28] \\
Clip judge on eight frames & 8 frames, photos & 0.80 & [0.66, 0.92] & +0.04 [$-$0.05, 0.15] \\
DINOv2 similarity & 3 frames, photos & 0.51 & [0.39, 0.65] & +0.34 [0.18, 0.48] \\
CLIP similarity & 3 frames, photos & 0.57 & [0.41, 0.72] & +0.27 [0.09, 0.47] \\
OCR text mismatch (EasyOCR) & 3 frames, photos & 0.57 & [0.49, 0.64] & +0.28 [0.14, 0.42] \\
OCR text mismatch (Tesseract, secondary) & 3 frames, photos & 0.58 & [0.50, 0.64] & +0.27 [0.14, 0.39] \\
DOVER (overall) & clip, no photos & 0.48 & [0.34, 0.63] & +0.37 [0.19, 0.56] \\
VideoScore2 (mean of 3 scores) & clip, text prompt & 0.57 & [0.44, 0.69] & +0.28 [0.12, 0.44] \\
Qwen3.5-9B, thinking on & 8 frames, photos & 0.67 & [0.49, 0.84] & +0.18 [0.04, 0.31] \\
Qwen3.5-9B, thinking on & clip, photos & 0.69 & [0.55, 0.82] & +0.15 [0.04, 0.26] \\
Qwen3.6-27B (FP8), thinking on & 8 frames, photos & 0.72 & [0.59, 0.86] & +0.12 [0.03, 0.22] \\
Qwen3.6-27B (FP8), thinking on & clip, photos & 0.67 & [0.53, 0.81] & +0.18 [0.06, 0.29] \\
Qwen3.6-27B (FP8), thinking off & 8 frames, photos & 0.73 & [0.60, 0.86] & +0.12 [$-$0.04, 0.26] \\
Qwen3.6-27B (FP8), thinking off & clip, photos & 0.68 & [0.57, 0.80] & +0.16 [0.01, 0.29] \\
Qwen3.5-9B, thinking off & 8 frames, photos & 0.54 & [0.39, 0.71] & +0.30 [0.15, 0.46] \\
Qwen3.5-9B, thinking off & clip, photos & 0.60 & [0.48, 0.74] & +0.25 [0.10, 0.38] \\
\bottomrule
\end{tabular}
\end{table}

\section{Label Sets Behind the Numbers}
\label{app:labels}

\Cref{tab:moreexperiments} lists two more experiments of \cref{sec:judges}, with their labels and answers. \Cref{tab:labelsets} lists each label set behind the judge, review and fix numbers.

\begin{table}[h]
\centering\small
\newcommand{\mk}[1]{\textsf{\textbf{#1}}}
\caption{Other judge experiments. Labels: \mk{B} listeners blind to production history, at least two per voice, \mk{L} one listener per line; numbers measure agreement with that reference only. Each choice was made and scored on the same labels, so numbers are optimistic.}
\label{tab:moreexperiments}
\setlength{\tabcolsep}{4pt}
\renewcommand{\arraystretch}{1.1}
\newcommand{\ax}[1]{\textbf{#1.}\hspace{0.3em}}
\begin{tabularx}{\textwidth}{@{}>{\raggedright\arraybackslash}p{4.2cm} >{\sffamily\bfseries}c@{\hspace{3.5pt}} >{\raggedright\arraybackslash}p{4.0cm} >{\raggedright\arraybackslash}X@{}}
\toprule
\hd{Axis and question} & \multicolumn{2}{l}{\hd{Labels}} & \hd{Answer and key numbers} \\
\midrule
\zebra \ax{Pronunciation} Which speech-to-text model; recheck on audio? & L & 208 lines: audio only (blind), then with script and judge output & \textbf{Error-keeping, text-only recheck.} On the labels made after output was shown (54 defect lines, against 43 on the blind pass of \cref{tab:validity}), Gemini 3.5 Transcribe~\cite{google2026gemini35transcribe} and ElevenLabs Scribe v2~\cite{elevenlabs2026scribev2} returned the script word for word on 30\% and 46\% of defect lines, against 6\% for ours, Azure Speech~\cite{microsoft2026azurespeech}; MAI-Transcribe-2~\cite{microsoft2026maitranscribe2} also caught fewer defects than ours. Audio in the recheck cut recall from 44 to 38 of 54 \\
\ax{Sound} Which text-to-speech voices to allow? & B & 107 voices on an earlier engine, three ad reads each; four listeners & \textbf{65 of 107,} excluding voices with poor recording quality, mispronunciations, unnatural intonation or an accent that did not fit the target market. Each voice was heard by at least two listeners, and the 44 split voices went to a tiebreak with earlier votes hidden; two voices users had swapped away from in production were among the 42 excluded, and the three most-cast voices were kept. The engine that narrated the round was admitted with all 30 of its voices \\
\bottomrule
\end{tabularx}
\end{table}

\begin{table}[h]
\centering\small
\caption{Label sets. ``Blind'' means the labelers saw no judge output. Assisted labels were made with judge findings on screen, and a thumbs up wrote the finding into the answer.}
\label{tab:labelsets}
\begin{tabularx}{\textwidth}{@{}>{\raggedright\arraybackslash}p{4.2cm} >{\raggedright\arraybackslash}p{2.7cm} >{\raggedright\arraybackslash}p{3.8cm} >{\raggedright\arraybackslash}p{1.9cm} >{\raggedright\arraybackslash}X@{}}
\toprule
\hd{Label set} & \hd{Size} & \hd{Labelers} & \hd{Blind?} & \hd{Used for} \\
\midrule
Annotator product labels on production clips & 120 decided scenes of 146 labeled, 29 defective & four annotator accounts; of the 120 decided scenes, 105 have two votes, ten have three or four, five have one; they miss subtle defects & yes & product judge and baselines (\cref{tab:experiments}, \cref{tab:baselines}) \\
\zebra Earlier review round & 66 accepted scene findings & reviewer thumbs & assisted & what scene judges flag \\
Scene judges, 120-ad round & 489 reviewer-scene pairs & portal reviewers & assisted & one scene judge or two (\cref{tab:experiments}) \\
\zebra Pronunciation corpus & 208 narrated lines, earlier narration engine & five listeners, one per line, ten lines heard by all & first pass yes, second pass after reveal & recall and precision on the first pass (\cref{tab:validity}); speech-to-text model and recheck on the second (\cref{tab:moreexperiments}) \\
Fresh pronunciation sample & 100 production lines (70 flagged, 20 near misses, 10 passed) & one listener & yes & precision on the 70 flagged lines: 0.21 [0.12, 0.31] \\
\zebra Talking-over clips & 50 of 128 clips & stem-separation loudness & no human labels & sound (\cref{tab:moreexperiments}) \\
Script judge study & 155 ads, 200 ratings & four raters, shown both judge versions' calls & no & reading listing photos \\
\zebra Caption placement & 100 captions & models against themselves & no human labels & placement pick stability (\cref{tab:experiments}) \\
Caption marks, 120-ad round & 530 captions & portal reviewers & yes, no caption judge shown & caption judges, run after the round; the placement prompt was chosen on these marks (\cref{tab:experiments,tab:validity}) \\
\zebra 120-ad review round & 562 scenes, 278 reviews & five internal reviewers & assisted & split, time, agreement (\cref{sec:portal}) \\
Reviewers alone & 60 of the round's ads, 280 scenes & two reviewers, R1 and R2, 60 ads each & R1 yes; R2 had seen judge findings in the round & the same ads with judges hidden (\cref{app:alone}) \\
\zebra Repair test & 30 scenes with garbled product text, 26 rated & one internal reviewer & yes & Repair (\cref{sec:fixing}) \\
Respelling test & 75 flagged lines, two takes each & pronunciation judge only & no human labels & respelling against a plain second take (\cref{app:gallery}) \\
\zebra Quoted-line rewrite & 129 scripts the judge cleared after the rewrite, of 134 it flagged in 410 earlier ads & two annotators, then an adjudicator; no controls & first verdict yes & regenerating only the quoted lines (\cref{sec:fixing}) \\
Voice listening study & 107 voices on an earlier narration engine, three ad reads each & four listeners, at least two per voice; split voices to a tiebreak & yes (blind to production history) & voices to allow (\cref{tab:moreexperiments}) \\
\bottomrule
\end{tabularx}
\end{table}

\section{Models}
\label{app:models}

The judges whose findings reviewers saw in the round run on closed models: Gemini Flash for the product clip judge, the scene clip judge, the sound judge and the listing-photo text reader; a GPT-5.6 model for the scene frames judge; Azure Speech with a GPT-5.6 model for pronunciation; and Claude Sonnet for the script judge. The caption judges, run only after the round, use Gemini Flash on the whole rendered ad. Gemini names are aliases that the provider updates, so a rerun may reach a newer model.

The edit techniques of \cref{tab:fixes} can run on several kinds of models. Repair needs a generative upscaler that regenerates the scene's own clip at higher resolution (video-to-video), such as SeedVR2~\cite{wang2026seedvr2}, STAR~\cite{xie2025star} or Topaz Starlight~\cite{topazlabs2025starlight}. A new scene needs a reference-to-video model, and a clip edit a video-editing model such as Gemini Omni, Seedance or Kling. Lines can be voiced with text-to-speech voices such as Gemini's or ElevenLabs', and a whole ad re-voiced speech-to-speech. Line suggestions can come from a language model, and music from a text-to-music model such as Lyria or Eleven Music. A logo is cut out from its background by a background-removal model. At generation, the quoted-line rewrite pairs the script judge with a language-model writer. Moving or restyling a caption uses a vision-language model and rewording one a language model (\cref{tab:fixes}); borrowing a scene and moving a logo call no model.

\section{Review Rubric}
\label{app:rubric}

\Cref{tab:rubric} lists the questions of the review rubric as they were used in the 120-ad round (\cref{sec:portal}).  The scene-quality grid asked for one answer on every rateable scene, meaning a scene with a generated clip or an animated still and never the end card; for most of the round the grid skipped scenes already marked product wrong and counted them as swaps.

\begin{table}[h]
\centering
\footnotesize
\caption{Questions of the review rubric in the 120-ad round, in the order the portal showed them in the round. The last answer of each question is the problem answer; it opens the picker and the secondary options. The two overlay-text rows are the two parts of the captions axis of \cref{tab:axes}.}
\label{tab:rubric}
\setlength{\tabcolsep}{4pt}
\begin{tabularx}{\textwidth}{@{}>{\raggedright\arraybackslash}p{2.6cm} >{\raggedright\arraybackslash}p{3.4cm} >{\raggedright\arraybackslash}p{3.6cm} X@{}}
\toprule
\hd{Question} & \hd{Answers} & \hd{A problem answer points to} & \hd{Secondary options} \\
\midrule
The product & Product looks right; Product wrong & every scene where it happens & Wrong product; Shape, color, logo or markings wrong; AI issue; Product text not intact, graded Low (``still readable'') or High (``gibberish''); a comment per scene \\
\zebra Scene quality & No opinion; Keep; Swap out & each rateable scene, which gets its own answer & AI issue; Repetitive, dull or crowded scene; World or continuity breaks; Something moves impossibly; Weak ending (last scene only); free-text boxes for another reason and for why \\
Pronunciation & Sounds right; Mispronounced & every spoken line where it happens & none; an optional box for the word \\
\zebra Sound quality & Sounds right; Issues & the whole ad & ``Which part?'', at least one of The voice and The music; an optional note \\
Script quality & Good; Bad & every spoken line where it happens & Factually incorrect (an unsupported claim); Awkward or unnatural phrasing; an optional note \\
\zebra Overlay text: placement & Correct; Can be better & every captioned scene where it happens & none \\
Overlay text: relevance & Correct; Issues & every captioned scene where it happens & none \\
\bottomrule
\end{tabularx}
\end{table}

\end{document}